\documentclass[10pt,twocolumn,letterpaper]{article}

\usepackage[pagenumbers]{iccv}              

\usepackage{tabularx}
\usepackage{tocloft}
\usepackage{multibib}
\usepackage{color, colortbl}
\usepackage[dvipsnames]{xcolor}
\newcommand{\red}[1]{{\color{red}#1}}
\newcommand{\blue}[1]{{\color{blue}#1}}

\newcommand{\methodname}{PHD}

\usepackage{mathtools}

\makeatletter
\renewcommand\paragraph{\@startsection{paragraph}{4}{\z@}%
                {2ex plus 0.2ex minus 0.1ex}%
                {-1.0em}%
                {\reset@font\normalsize\bfseries}}

\definecolor{green}{rgb}{0.0, 0.5, 0.0}

\definecolor{iccvblue}{rgb}{0.21,0.49,0.74}
\usepackage[pagebackref,breaklinks,colorlinks,allcolors=iccvblue]{hyperref}

\def\paperID{2038} 
\def\confName{ICCV}
\def\confYear{2025}

\title{PHD: \underline{P}ersonalized 3D \underline{H}uman Body Fitting with Point \underline{D}iffusion}

\author{
~~~~~~~~~~~~~~~~~~~~~~~~~~~~~~~~~~~~~~~~~~~~~Hsuan-I Ho$^{1,3}$
\and
Chen Guo$^{1,3}$~~~~~~~~~~~~~~~~~~~~~~~~~~~~~~~~~~~~~~~~~~~~~
\and
Po-Chen Wu$^{3}$
\and
Ivan Shugurov$^{3}$
\and
Chengcheng Tang$^{3}$
\and
Abhay Mittal$^{3}$
\and
Sizhe An$^{3}$
\and
~~~~~~~~~~~~~~~~~~~~~~~~~~~~~~Manuel Kaufmann$\dagger^{2}$
\and
Linguang Zhang$\dagger^{3}$~~~~~~~~~~~~~~~~~~~~~~~~~~~~~~
\and
\\
$^{1}$Department of Computer Science, ETH Zürich \\
$^{2}$ETH AI Center, ETH Zürich \\
$^{3}$Reality Labs, Meta
}

\begin{document}
\twocolumn[{%
\renewcommand\twocolumn[1][]{#1}%
\maketitle
\begin{center}
    \centering
    \captionsetup{type=figure}
    \includegraphics[width=\textwidth,trim={0 2cm 0 0},clip]{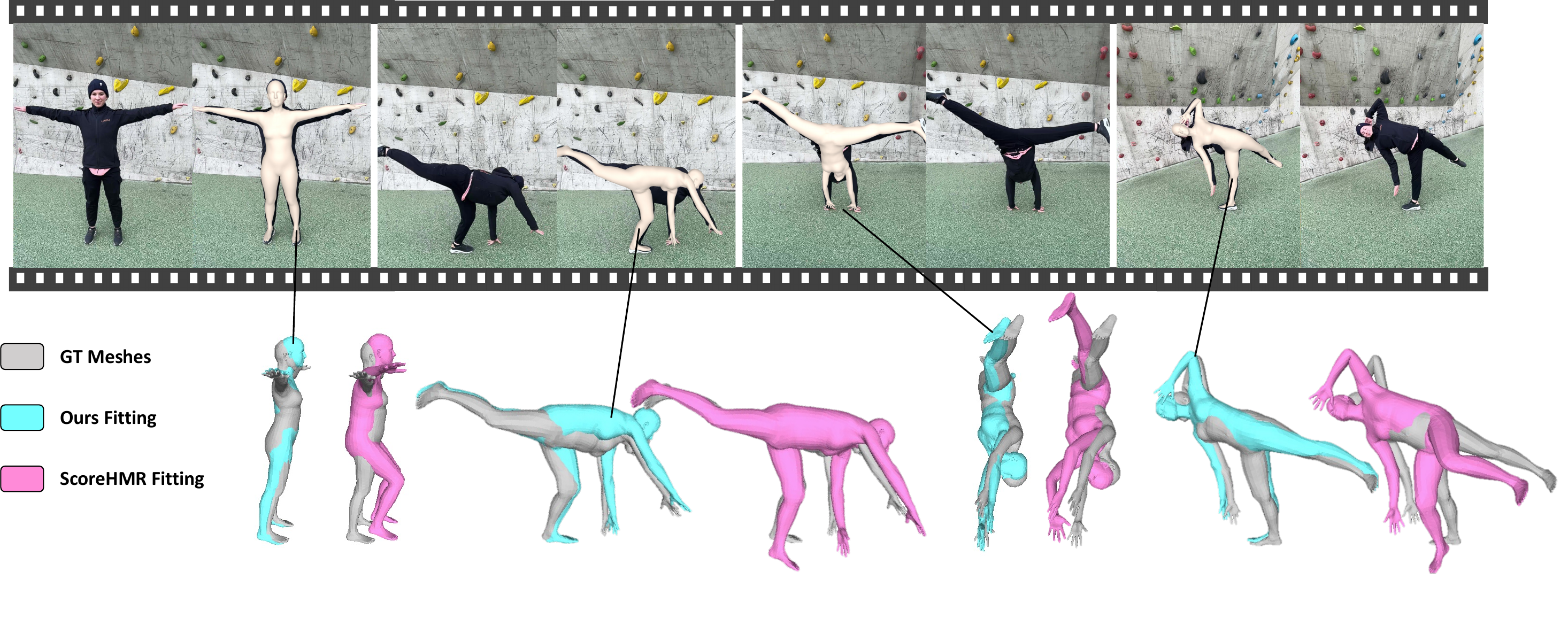}
    \caption{We present~\methodname, a novel body fitting paradigm for obtaining accurate 3D body poses from single videos.
    \methodname~handles complex poses well and avoids common problems of existing methods, like bending knees, that are due to inaccurate shape estimates.
    }
    \label{fig:teaser}
\end{center}%
}]
\def\thefootnote{$\dagger$}\footnotetext{Equal advisory}

\def\thefootnote{\arabic{footnote}}

\begin{abstract}
We introduce PHD, a novel approach for personalized 3D human mesh recovery (HMR) and body fitting that leverages user-specific shape information to improve pose estimation accuracy from videos.
Traditional HMR methods are designed to be user-agnostic and optimized for generalization. While these methods often refine poses using constraints derived from the 2D image to improve alignment, this process compromises 3D accuracy by failing to jointly account for person-specific body shapes and the plausibility of 3D poses.
In contrast, our pipeline decouples this process by first calibrating the user's body shape and then employing a personalized pose fitting process conditioned on that shape.
To achieve this, we develop a body shape-conditioned 3D pose prior, implemented as a Point Diffusion Transformer, which iteratively guides the pose fitting via a Point Distillation Sampling loss.
This learned 3D pose prior effectively mitigates errors arising from an over-reliance on 2D constraints.
Consequently, our approach improves not only pelvis-aligned pose accuracy but also absolute pose accuracy -- an important metric often overlooked by prior work.
Furthermore, our method is highly data-efficient, requiring only synthetic data for training, and serves as a versatile plug-and-play module that can be seamlessly integrated with existing 3D pose estimators to enhance their performance.
Code and models are available at
\url{https://PHD-Pose.github.io}.
\end{abstract}
\vspace{-1.2em}    
\section{Introduction}
\label{sec:intro}
Perceiving accurate 3D human pose and shape is fundamental for future AI systems, such as realistic human avatars for AR/VR telepresence, personalized service robots, and human behavior understanding.
Impressive strides have recently been made in 3D human pose and shape estimation~\cite{Bogo:ECCV:2016, kanazawa2018end, goel2023humans, stathopoulos2024score, patel2025camerahmr, sarandi2024nlf, kolotouros2019spin, kocabas2019vibe}.
Most of these methods are designed to be subject-agnostic, focusing on broad generalization, and are trained on large corpora of 3D pseudo-labels relying on 2D reprojection objectives. We identify two main problems in existing work.

First, they miss a key opportunity to enhance 3D pose accuracy by leveraging the consistency of user identity over time. Typically, these methods estimate body shape, pose, and pelvis position (in camera coordinates) simultaneously per frame.
However, the shape parameters they predict often vary across frames, contradicting the assumption that a subject's body shape does not change in a short video.
This entanglement of pose and shape can lead to undesirable situations where changes in pose must make up for shape errors to satisfy the optimization objective.
Second, existing methods tend to overly rely on 2D objectives to achieve pose-to-image alignment.
This often comes at the expense of 3D accuracy.
This problem affects optimization-based methods, but also regressors that are trained on 3D pseudo-data, which has been obtained by fitting initial 3D pose estimates with 2D objectives.

\begin{figure}[t]
\centering
\includegraphics[width=\linewidth]{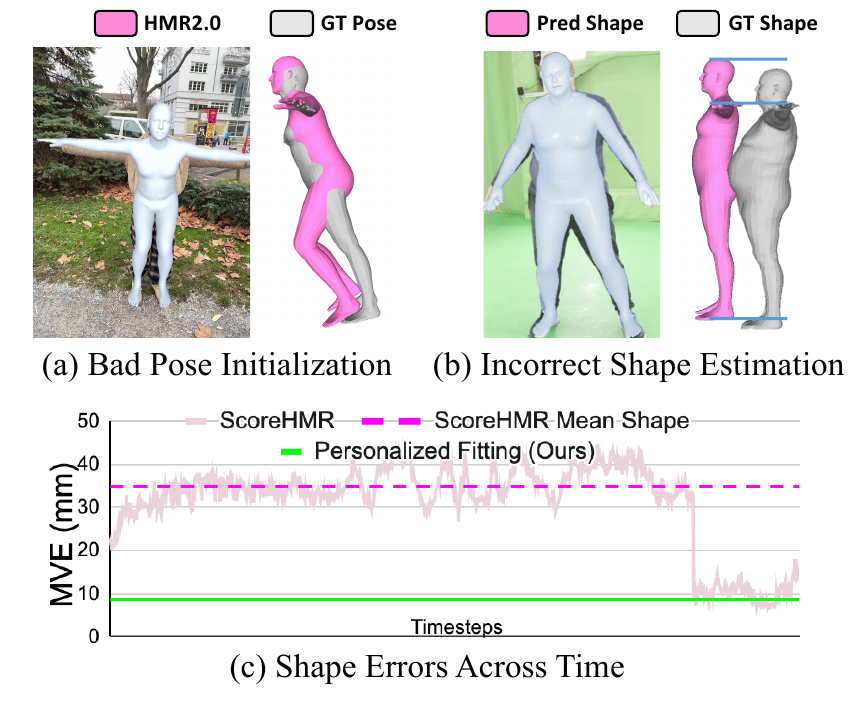}
\caption{
\textbf{Shape estimation issues in existing methods.} (a) Incorrect shape estimations result in unrealistic bending poses to satisfy 2D projections. (b) Regression-based methods often fail to accurately estimate shapes from a single image. (c) Shape estimations often vary over time, making temporal smoothness challenging.
Our personalized fitting method estimates a fixed, more accurate shape and mitigates these problems.
}
\label{fig:personalization}
\end{figure}

As a result of these two problems, existing body fitting pipelines often fail to simultaneously ensure the accuracy of body shape, body pose, and pelvis position (see Fig.~\ref{fig:personalization}). In this work, we address these limitations through an innovative \emph{personalized} 3D body pose estimation method, as opposed to existing \emph{generalized} methods.

We demonstrate that explicitly considering user-specific information is essential for achieving consistently accurate body pose estimation over time.
Our approach decomposes the body pose estimation problem into two stages: a shape estimation stage, referred to as personalization, and a pose fitting stage that uses the shape as a conditional input.
To perform personalization, we introduce \textbf{SHAPify}, a body shape fitting method that calibrates a user's shape parameters from a single video frame showing a reference pose and optionally uses as little as the height and weight of the person as additional identity information.
SHAPify provides highly accurate shape estimations compared to existing generalized pose and shape estimators.

The subsequent pose fitting stage is inspired by existing methods~\cite{Bogo:ECCV:2016,kolotouros2019spin,song2020humanbodymodelfitting,kolotouros2021prohmr,stathopoulos2024score} that follow a ``regress-then-refine" scheme. These methods begin by jointly estimating shape, pose, and pelvis position using a regressor, followed by an optimization process that aims to minimize 2D alignment errors using 2D keypoint constraints. 
Although widely used in pose refinement~\cite{Bogo:ECCV:2016,kolotouros2019spin,song2020humanbodymodelfitting,kolotouros2021prohmr,stathopoulos2024score} and avatar reconstruction~\cite{jiang2022instantavatar, guo2023vid2avatar, moon2024exavatar, guo2025vid2avatarpro}, this approach is limited.
A strong reliance on 2D objectives is problematic as they may easily degrade 3D pose accuracy due to the neglect of depth information (\eg, see~\cref{fig:ablation_point}).
Furthermore, existing methods are often at the mercy of the provided initializations, \ie, they often cannot recover from a bad initialization.
We argue that a good 3D prior should be able to do so, \ie, improve the bad parts of the initialization while retaining the good ones.

These limitations motivate us to incorporate a novel, strong 3D pose prior into the ``regress-then-refine'' fitting process, called \textbf{PointDiT}.
The 3D pose prior encourages the pose output to remain within the manifold of plausible poses, preventing overfitting to 2D observations.
At the same time, it is less reliant on good initializations than previous work and can correct bad initializations.
More specifically, we formulate the prior as a body shape-conditioned point diffusion model that learns to sample body surface points.
We demonstrate that the use of surface points as the pose representation more effectively reconstructs uncommon poses compared to the use of joint angle-based counterparts~\cite{stathopoulos2024score, Cho_2023_ICCVW}.
PointDiT is then iteratively applied to guide the pose fitting process via a newly designed Point Distillation Sampling loss.
By integrating the 3D pose prior into the fitting loop, we effectively constrain the plausibility of 3D poses when optimizing with 2D alignment objectives.
It is worth noting that PointDiT is efficiently trained only with the synthetic dataset BEDLAM~\cite{Black_CVPR_2023}, and can be easily applied to off-the-shelf 3D pose estimators as a versatile plug-and-play module.
In our experiments, we demonstrate further accuracy improvements over our closest related work ScoreHMR~\cite{stathopoulos2024score}, and establish a new state of the art on EMDB~\cite{kaufmann2023emdb} for both pelvis-aligned and \emph{absolute} pose accuracy -- a metric often underemphasized in prior work. In summary, our contributions are as follows:
\begin{itemize}
    \item A new personalized body pose estimation paradigm, \methodname, that utilizes user-specific shape information to improve pose accuracy by decoupling the estimation of body shape, pose, and pelvis position.
    \item A robust body shape-conditioned 3D body pose prior based on a point diffusion model, PointDiT, which we use with a Point Distillation Sampling loss to provide effective 3D pose guidance in personalized body fitting. 
    \item We demonstrate that \methodname~is versatile, accurate, and improves pose initializations supplied by strong baselines.
\end{itemize}

\section{Related Work}
\label{sec:work}
\paragraph{3D Human pose estimation and body fitting.}
Estimating 3D human pose and shape from monocular inputs has been extensively studied, ranging from optimization-based approaches to recent transformer-based regressors.
Early optimization-based methods~\cite{Robertini_2016, Habermann_livecap, xu_monoperfcap} performed 3D pose tracking by fitting pre-scanned body templates to video sequences. However, obtaining such templates demands specialized capturing equipment and is not easy to scale up.
Therefore, later work resorts to fitting a parametric body model~\cite{SMPL:2015, pavlakos2019expressive, Xu_2020_CVPR, Joo_2018_CVPR} to 2D observations, such as landmarks~\cite{Bogo:ECCV:2016, pavlakos2019expressive, Xu_2020_CVPR}, masks~\cite{omran2018neural}, and body part segmentations~\cite{lassner2017unite}.
While optimization-based methods achieve promising fitting results, they often involve lengthy optimization loops and are prone to overfitting 2D objectives. 
Learning-based approaches, instead, directly regress the parametric body model from images~\cite{kanazawa2018end, kolotouros2019spin, Kocabas_PARE_2021, li2021hybrik, DSR:ICCV:2021, ROMP, BEV, cai2023smplerx, lin2023one, sun2024aios, sarandi2024nlf}, videos~\cite{kocabas2019vibe, kanazawa2019learning, TRACE, shin2023wham}, or estimate human bodies as mesh vertices~\cite{lin2021mesh, kolotouros2019convolutional, sarandi2020metric, cho2022cross, corona2022learned}. Recent advancements successfully employed transformer-based architectures~\cite{dosovitskiy2020vit, xu2022vitpose} for learning-based 3D pose estimation and achieved state-of-the-art performance \cite{goel2023humans, dwivedi_cvpr2024_tokenhmr, saleem2024genhmrgenerativehumanmesh, patel2025camerahmr, cai2023smplerx, sun2024aios}. To further enhance the capability of transformer models, 3D pseudo ground-truth annotations~\cite{joo2021exemplar, moon2022neuralannot, goel2023humans, li2022cliff} are widely used for training. While this ensures better 2D pose-to-image alignment, their 3D pose plausibility is reduced. 

Recently, a new line of work~\cite{song2020humanbodymodelfitting, wang23refit, stathopoulos2024score, ludwig2024leveraging} combines both methods by first estimating the initial human body and then further refining it. The recent ScoreHMR \cite{stathopoulos2024score} is our closest related work. Note that these methods are all generalized models designed to be subject-agnostic. Different from the generalized methods, our goal is to leverage only minimal personal information~\cite{ludwig2024leveraging} (i.e., heights and weights) to achieve more accurate and personalized pose and shape estimation.

\paragraph{3D human pose prior.}
3D pose priors are essential for numerous tasks, \eg, fitting 3D human poses to images and videos \cite{rempe2021humor, dposer, Bogo:ECCV:2016, pavlakos2019expressive, tiwari22posendf}, modeling pose ambiguity~\cite{sengupta2023humaniflow, dwivedi2024poco}, and inpainting body poses~\cite{kaufmann2020convolutional, ho2021render}. 
Existing pose priors can be broadly categorized into two types: unconditional and conditional ones.
For instance, early unconditional pose priors focused on learning joint limits~\cite{akhter2015pose} to avoid implausible poses.
Recently, pose priors leveraging Gaussian Mixture Models~\cite{Bogo:ECCV:2016}, Generative Adversarial Networks~\cite{georgakis2020hierarchical, kanazawa2018end}, VAEs~\cite{pavlakos2019expressive, dwivedi_cvpr2024_tokenhmr}, and implicit neural functions~\cite{tiwari22posendf, he24nrdf}, are employed to impose unconditioned pose priors for model training and fitting. However, many of these unconditional priors are biased towards common training samples and do not generalize well to unseen poses. 
Of late, the community has embraced conditional prior models, \eg, ProHMR~\cite{kolotouros2021prohmr}, GenHMR~\cite{saleem2024genhmrgenerativehumanmesh}, and ScoreHMR~\cite{stathopoulos2024score}, where the prior is trained to condition on an input image. Training with more diverse visual data, these methods demonstrated better generality to unseen images and poses.

Building on the foundations of previous conditional prior models, we develop a generative model that captures the conditional distribution of body poses given an input image \textit{and} the human body shape.
Previously, ScoreHMR and Cho~\etal \cite{Cho_2023_ICCVW} used angular parameters to represent body poses during training. Our experiments (\cref{sec:ablation}) suggest that this approach is suboptimal for image and shape conditioning, due to the weak correlation between angular parameters and conditional inputs.
Inspired by related work that has recognized denser representations \cite{guler2018densepose,Zhang:MOJO,sarandi2024nlf}, we instead build our prior with 3D points sampled from the body surface.
This is a natural choice given that we condition on a 3D body shape, where 3D body points are more closely related to than angular parameters are.
We integrated this novel point-based pose prior into the pose fitting process for robust personalized 3D pose estimation. 
\begin{figure*}[t]
\centering
\includegraphics[width=\linewidth]{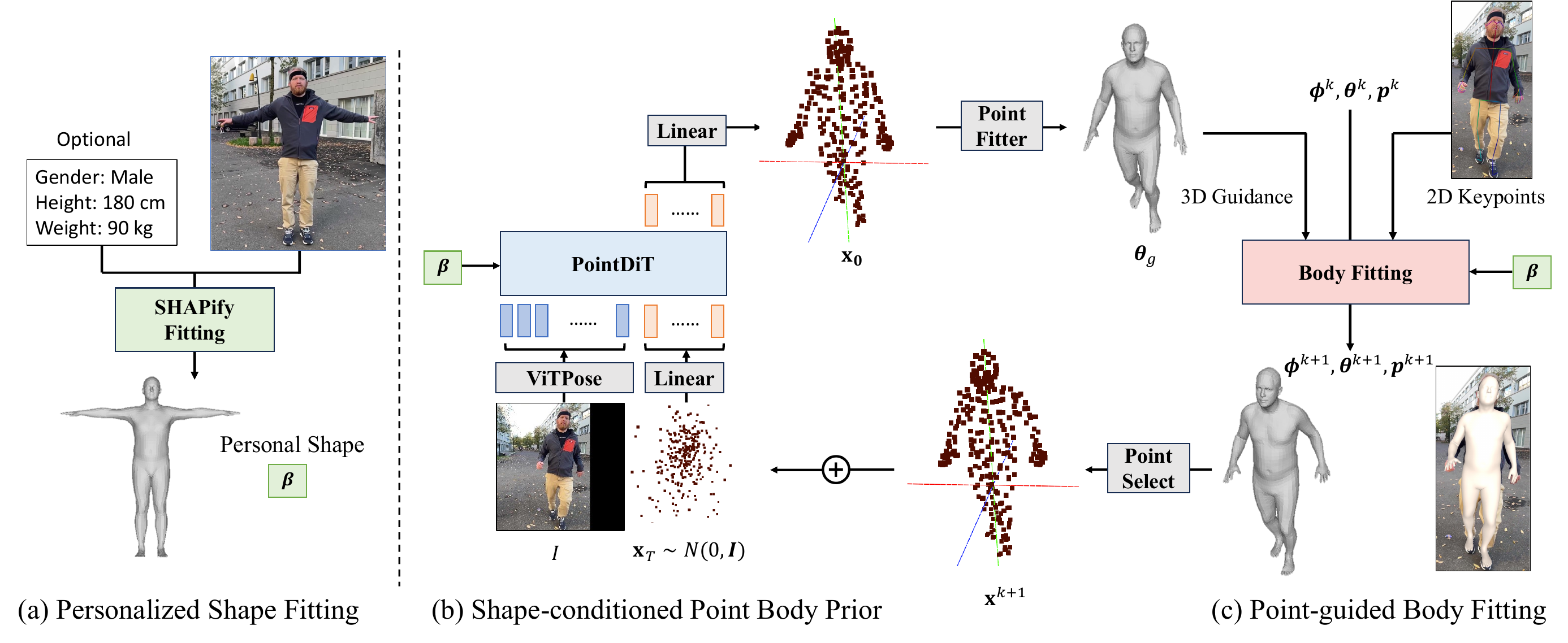}
\caption{\textbf{Method overview}. Our personalized body fitting pipeline consists of two major steps. (a) In the personalization step, we estimate personal shape information $\beta$ from a single RGB image and optional body measurement information. (b) This personal shape parameter, $\beta$, is used by the learned point body prior, PointDiT, allowing for the sampling of body points given the shape and image conditions. These body points are used for guiding the body fitting process. (c) In the body fitting process, we alternately refine the 3D pose parameters $\phi$, $\theta$, and $\mathbf{p}$, and sample new body points $\mathbf{x}_0$ to ensure that the body poses comply with both the 2D projection and the learned 3D distribution.
}
\label{fig:framework}
\end{figure*}

\section{Methodology}
\label{sec:method}
\paragraph{Preliminaries.}
\label{sec:preliminaries}
We use the SMPL~\cite{SMPL:2015} parametric body model to represent 3D body shape and pose. The body shape is parameterized by PCA coefficients $\beta \in \mathbb{R}^{10}$, which SMPL maps to a person-specific mesh in the reference T-pose. The body pose is composed of 23 relative joint rotations $\theta \in \mathbb{R}^{23\times3}$ using the axis-angle representation, along with pelvis orientation $\phi \in \mathbb{R}^{3}$ in the camera coordinate system. Finally, we additionally define $\mathbf{p} \in \mathbb{R}^{3}$ as the relative position of SMPL pelvis to the camera center (origin).

\paragraph{Method overview.}
\label{sec:overview}
Our goal is to estimate the shape parameters $\beta$, bone rotations $\theta$, per-frame pelvis orientation $\phi$, and location $\mathbf{p}$ from a monocular RGB video of a moving human, given the camera's focal length. 
Our personalized body fitting pipeline comprises two main steps, as illustrated in ~\cref{fig:framework}.
First, we extract person-specific shape information by estimating the shape parameters from a calibration frame where the person is approximately in a rest pose (\eg, T-pose or I-pose). (\cref{sec:shape_fitting}). This step is performed only once for each subject. 
In the second body fitting step, a point diffusion model (\cref{sec:point_dit}) that learns to sample body points conditioned on the image and shape parameters is trained to serve as a 3D prior for optimizing the body pose. 
We then iteratively sample body points and refine the pose parameters by considering both 2D keypoint projections and 3D prior guidance (\cref{sec:fitting}). 

\subsection{Personalized Body Shape Estimation}
\label{sec:shape_fitting}

Obtaining accurate body shape from a single image is an ill-posed problem due to occlusions by clothing and scale/pose ambiguity, which poses challenges for existing data-driven methods. Inspired by SHAPY~\cite{choutas2022accurate} and SMPLify~\cite{Bogo:ECCV:2016}, we develop an optimization-based method, \textbf{SHAPify}, to reliably obtain shape parameters from a single calibration image, 2D keypoints obtained from an off-the-shelf detector, and optional body measurements (see~\cref{fig:framework} (a)).

\paragraph{Shape calibration setup.}
To estimate an accurate body shape from a single image, we capture the subject in a rest pose. In the standard 3D pose and shape estimation problem, shape parameters $\beta$, bone rotations $\theta$, pelvis orientation $\phi$, and position $\mathbf{p}$ are all unknown variables.
We initialize $\theta$ and $\mathbf{p}$ using the predefined rest pose and prioritize changes in $\phi$ and $\beta$ over changes in $\theta$ and $\mathbf{p}$ during optimization by adjusting the learning rates.

\paragraph{Optimization objectives.}
We minimize the following objective:
\begin{equation}
    \mathcal{L}_\text{rep} = \Vert \Pi(\mathbf{J_{3D}(\phi,\theta,\beta)} + \mathbf{p}) - \mathbf{J_{2D}}\Vert_{1},
    \label{eq:2d_rep}
\end{equation}
where $\mathbf{J_{3D}}$ is the SMPL differentiable LBS function for obtaining 3D joints, $\Pi(\cdot)$ is the 3D-to-2D projection, and $\mathbf{J_{2D}}$ is the 2D keypoints obtained from a keypoint detector.
We choose smaller learning rates for $\theta$ and $\mathbf{p}$ to encourage changes in shape parameters $\beta$ and pelvis orientation $\phi$.

2D keypoints alone do not provide sufficient constraints on the body shape, as the keypoint projection depends on both the pelvis' pitch angle and $\beta$. To mitigate this ambiguity, we impose additional regularization terms to ensure the human shape follows specific body measurements:
\begin{equation}
    \mathcal{L}_\text{reg} = \lambda_\beta\Vert \beta \Vert^{2}_{2}+ \lambda_h\Vert H(\beta) - h\Vert_{1} + \lambda_w\Vert W(\beta) - w\Vert_{1}.
\label{eq:reg}
\end{equation}
Here, $(h, w)$ are average human height and weight, which can be replaced by user-provided measurements if available, and $H(\cdot), W(\cdot)$ are differentiable functions that compute the height and weight from a SMPL body mesh (see Supp. Mat.~\cref{sec:shapify_details}). The final objective then is:
\begin{equation}
    \mathrm{argmin}_{\phi,\theta,\beta,\mathbf{p}} \mathcal{L}_\text{rep}(\phi,\theta,\beta,\mathbf{p})+\mathcal{L}_\text{reg}(\beta).
\label{eq:shape_all}
\end{equation}

\subsection{Shape-conditioned Point Diffusion Priors}
\label{sec:point_dit}

\paragraph{Diffusion model and rectified flow.}
Diffusion models iteratively perturb and denoise data samples across several diffusion steps $T$ \cite{Ho2020DiffusionModels}. Given a sample $\mathbf{x}_0$, the forward process gradually blends the sample with Gaussian noise $\epsilon \sim \mathcal{N}(0, \mathbf{I})$ to obtain noisy samples $\mathbf{x}_t$.
To sample new data, a reverse process is applied to recover clean data from random noise, whereby the noise $\mathbf{\hat{\epsilon}}(\mathbf{x}_t, t)$ at step $t$ is estimated by a neural network.
However, diffusion models typically require $T > 20$ denoising steps for optimal performance, which is expensive for a pose fitting procedure.
Recently, several works~\cite {liu2022flow,sd3} incorporated the rectified flow formulation into diffusion models to reduce the number of denoising steps. Here, the forward process is considered a simple linear interpolation between sample $\mathbf{x}_0$ and noise $\mathbf{\epsilon}$:
\begin{equation}
    \mathbf{x}_t = (1-\frac{t}{T})\mathbf{x}_0+\frac{t}{T}\mathbf{\epsilon}.
\label{eq:flow_forward}
\end{equation}
By replacing the $t$ variable, we can obtain $\mathbf{x}_t = \mathbf{x}_{t-1} + \mathbf{u}$, where $\mathbf{u}=\epsilon-\mathbf{x}_0$ represents the ``flow". Similar to the standard diffusion model, a reverse process for sampling can also be represented with a neural network:
\begin{equation}
    \mathbf{x}_{t-1} = \mathbf{x}_t - \mathbf{\hat{u}}(\mathbf{x}_t, t),
\label{eq:flow_backward}
\end{equation}
where $\mathbf{\hat{u}}(\mathbf{x}_t, t)$ predicts the flow $\mathbf{\hat{u}}=\hat{\epsilon}-\mathbf{\hat{x}}_0$ at time $t$ given $\mathbf{x}_{t}$.
The training loss follows conditional flow matching:
\begin{equation}
    \begin{aligned}
    \mathcal{L}_\text{CFM}(\mathbf{x})  &= \mathbb{E}_{t\sim \mathcal{U}(t),\epsilon\sim \mathcal{N}(0, \mathbf{I})} \Bigl[ \Vert  \mathbf{\hat{u}}(\mathbf{x}_t, t) - \mathbf{u} \Vert^{2}_{2} \Bigr] \\ 
    &= \mathbb{E}_{t\sim \mathcal{U}(t),\epsilon\sim \mathcal{N}(0, \mathbf{I})}  \Bigl[ w_t\Vert  \mathbf{\hat{\epsilon}}(\mathbf{x}_t, t) - \mathbf{\epsilon} \Vert^{2}_{2} \Bigr].
    \end{aligned}
\label{eq:cfm}
\end{equation}
We follow the reweighing and scheduling of Stable Diffusion 3~\cite{sd3} to train our model with the $\mathcal{L}_\text{CFM}$ loss.

\paragraph{Point diffusion transformer.}
Our objective is to develop a generative model that captures the conditional distribution of body poses~\cite{kolotouros2021prohmr,stathopoulos2024score} given an input image $I$ \emph{and} personalized body shape.
We call our novel human body prior \textbf{PointDiT} -- it is a 3D point diffusion model designed to sample body points.
The detailed architecture of PointDiT is illustrated in Supp. Mat.~\cref{sec:pointdit_details}. We adapt the original Diffusion Transformer architecture~\cite{peebles2023scalable} with the following modifications:
(1) We extract image tokens and 2D heatmaps from ViTPose~\cite{xu2022vitpose} to construct conditional tokens $\mathbf{c}^1, ..., \mathbf{c}^{256}$ for self-attention conditioning.
(2) We replace the class embedding with shape parameters $\beta$ in adaLN-Zero conditioning. 
(3) We adopt the rectified flow scheduling and re-weighting from SD3 \cite{sd3}. During training, noisy body points $\mathbf{x}_t$ are generated by perturbing ground-truth points $\mathbf{x}_0$ with random noises $\epsilon$, as described in \cref{eq:flow_forward}.
The flow-based formulation allows us to promptly sample point clouds in as few as $T=5$ denoising steps, which is crucial for integrating PointDiT into the fitting procedure. 

\paragraph{Model training.}
Our body point clouds are made up of $S = 238$ mesh vertices and $J = 45$ joints from the SMPL model.
We select $S$ based on the accuracy of the SMPL fitter in NLF~\cite{sarandi2024nlf}. 
The choice of $S$ and $J$ allows for accurate conversion of point clouds to SMPL pose parameters while still ensuring efficient training (see Supp. Mat.~\cref{fig:fitter}).
We train our model on the synthetic BEDLAM~\cite{Black_CVPR_2023} dataset as it provides high-quality images with ground-truth shape and pose parameters.
During each training iteration, we extract conditional image features $\mathbf{c}\in\mathbb{R}^{16\times16\times512}$ from the cropped $256 \times 256$ image $I$ and extract points $\mathbf{x}_0\in\mathbb{R}^{(S+J)\times3}$ from the ground-truth mesh. These points are linearly blended with Gaussian noise $\epsilon$ at a randomly sampled time step $\frac{t}{T} \in [0, 1)$, following \cref{eq:flow_forward}. The PointDiT model is then trained to predict the rectified flow $\mathbf{\hat{u}}(\mathbf{x}_t, t, \mathbf{c},\beta)$ at time step $t$, given $\mathbf{c},\beta$ as conditions, using the loss function in \cref{eq:cfm}.

\subsection{Prior-guided Body Fitting}
\label{sec:fitting}
\paragraph{Point distillation sampling.} 
 
We begin by detailing how PointDiT can serve as a pose prior for guiding the body fitting process. 
Inspired by the concept of Score Distillation Sampling~\cite{poole2022dreamfusion}, we introduce Point Distillation Sampling, an iterative process that leverages PointDiT to guide the fitting process.
As illustrated in~\cref{fig:framework} (b), a noisy point cloud is denoised using PointDiT to produce a clean point cloud $\mathbf{x}_0$, following the procedure outlined in~\cref{eq:flow_backward}.
Using the sampled point cloud, we compute two losses to enforce the 3D prior. The first loss is calculated as the pelvis-aligned L2 error between the sampled point clouds and the corresponding points derived from the fitted body parameters:
\begin{equation}
    \mathcal{L}_p = \lambda_p \Vert  \mathbf{x}_0 - \mathbf{P_{3D}}(\phi,\theta,\beta^*) \Vert _2.
    \label{eq:point}
\end{equation}
Here, $\beta^*$ denotes the calibrated shape parameters obtained from~\cref{sec:shape_fitting}, and $\mathbf{P_{3D}}$ is a differentiable Linear Blend Skinning function used to compute the selected points from SMPL parameters. 
The second loss, $\mathcal{L}_a$, uses the Point Fitter~\cite{sarandi2024nlf} to convert the point cloud $\mathbf{x}_0$ back to pose parameters $(\phi_{g},\theta_{g})$, and penalizes the difference:
\begin{equation}
\begin{aligned}
     \mathbf{x}_0 &\xmapsto{\text{Point Fitter}}  (\phi_{g}, \theta_{g} ) \\
    \mathcal{L}_a = \lambda_\phi \Vert \phi_{g} &- \phi  \Vert _2 +  \lambda_\theta \Vert \theta_{g} - \theta  \Vert _2,
\end{aligned}
\label{eq:angle}
\end{equation}
where $\lambda_{(.)}$ denotes the weights balancing each term. 

\paragraph{Sampling-fitting in the loop.}
The objective function of the fitting process, similar to existing methods, comprises a data term and a prior term:
\begin{equation}
    \mathrm{argmin}_{\phi,\theta,\mathbf{p}} \mathcal{L}_\text{data}(\phi,\theta,\beta^*,\mathbf{p})+\mathcal{L}_\text{prior}(\phi,\theta,\beta^*), 
    \label{eq:fitting_all}
\end{equation}
where the data term minimizes 2D keypoint errors:
\begin{equation}
    \mathcal{L}_\text{data} = \Vert \Pi(\mathbf{J_{3D}(\phi,\theta,\beta^*)} + \mathbf{p}) - \mathbf{J_{2D}}\Vert_{1},
    \label{eq:fitting_kp}
\end{equation}
The prior term is defined as $\mathcal{L}_\text{prior} = \mathcal{L}_{p} + \mathcal{L}_{a}$.
Unlike traditional fitting methods that use a fixed data prior throughout the optimization process, our approach incorporates a sampling-and-refinement loop to iteratively enhance the sampled point clouds for 3D guidance.
This is, at iteration $k$, the parameters $(\phi^k,\theta^k,\mathbf{p}^k)$ are updated based on~\cref{eq:fitting_all} (see~\cref{fig:framework} (c)). 
Then, $(\phi^{k+1},\theta^{k+1})$ are used to generate a new set of points: $\mathbf{x}^{k+1} = \mathbf{P_{3D}}(\phi^{k+1},\theta^{k+1})$. 
Subsequently, $\mathbf{x}^{k+1}$ is perturbed with a small noise level ($t/T = 0.75$) using~\cref{eq:flow_forward}, and a new point cloud is resampled for the next fitting iteration.

\begin{figure*}[t]
\centering
\includegraphics[width=\linewidth]{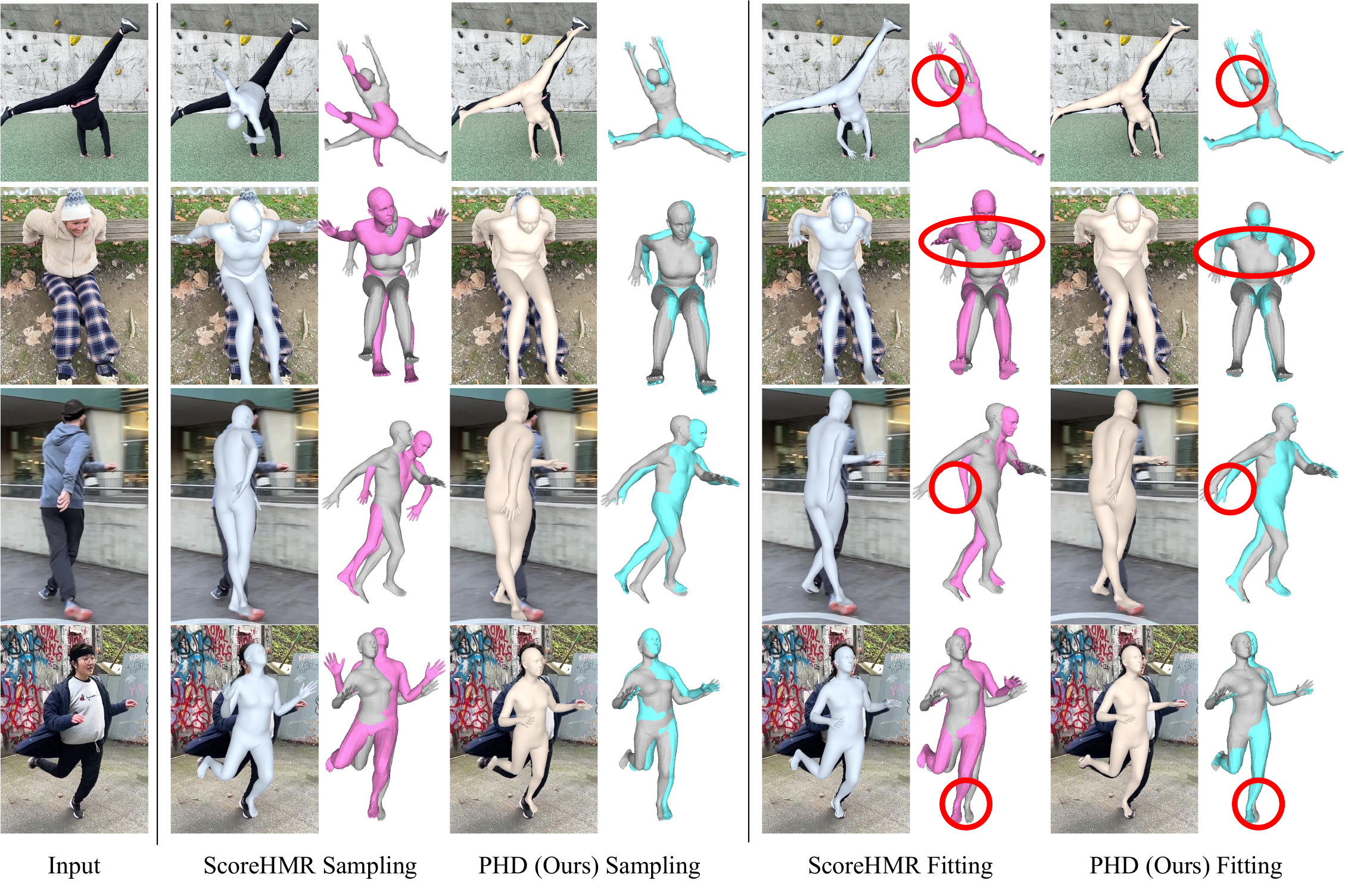}
\caption{\textbf{Qualitative comparison of body fitting}. \emph{Left}: Comparison when sampling poses in ScoreHMR and our method. Our method aligns better with both 2D images and 3D ground truth. \emph{Right}: Result of fitting HMR2.0b initializations with the ScoreHMR and our framework. ScoreHMR cannot recover from bad initializations as much as ours can. Ground-truth poses in gray. Please also refer to~\cref{tab:prior_res}}
\label{fig:big_fig}
\end{figure*}

\paragraph{Pose initialization.} Initialization is crucial in optimization problems. 
Unlike previous fitting methods that heavily rely on initializations from learned generalized regressors, we leverage PointDiT to sample plausible initial body poses.
We also show that our model is compatible out-of-the-box with most existing 3D pose regressors. 
For a comprehensive evaluation and discussion, please see~\cref{sec:compare_all}.

\section{Experiments}
\label{sec:exp}

\subsection{Experimental Setup}
\label{sec:exp_setup}
\paragraph{Dataset.}
\textbf{BEDLAM}~\cite{Black_CVPR_2023} is a synthetic dataset comprising 1M+ training images paired with ground-truth SMPL parameters. BEDLAM is the only dataset we use for training the shape-conditioned PointDiT model.
\textbf{EMDB}~\cite{kaufmann2023emdb} features in-the-wild videos of 10 subjects, with SMPL poses captured using precise electromagnetic sensors. It also includes accurate 3D global camera trajectories provided by the Apple AR Toolkit, along with ground-truth SMPL shapes derived from 3D scans. All the experiments and ablation studies are evaluated on the EMDB1 split.
In Supp. Mat., we also present results on the \textbf{3DPW}~\cite{von2018recovering} dataset, which captures daily life human performance with IMUs.

\paragraph{Evaluation protocol.}
In the 3D pose estimation literature, the accuracy of pelvis-aligned (local) poses is typically evaluated using the \textbf{MPJPE(-PA)} and \textbf{MVE(-PA)} metrics. However, these metrics do not consider pelvis position error, which is crucial for determining absolute pose accuracy in the camera coordinate system.
Therefore, we also report the \textbf{Pelvis Error}, which measures the distance between the predicted and ground-truth pelvis positions. Following the approach in SPEC~\cite{Kocabas_SPEC_2021}, we additionally report the absolute MPJPE metric in the camera coordinate system, referred to as \textbf{C-MPJPE}. To evaluate shape prediction accuracy, we report \textbf{per-vertex errors} of body meshes in the rest pose.

\subsection{3D Pose and Shape Accuracy}
\label{sec:compare_all}

\paragraph{Pelvis-aligned pose accuracy.} 
The problem setting of \textbf{ScoreHMR}~\cite{stathopoulos2024score} closely aligns with our method.
Therefore, we conduct a thorough comparison by providing ScoreHMR with identical inputs and the ground-truth focal length as used in our approach.
The quantitative results are presented in~\cref{tab:vsemdb}. We evaluate the methods using three different pose initialization strategies: 
(1) initializing by sampling a pose using the learned priors (referred to as \textbf{Sample init.}),
(2) initializing with HMR2.0b predictions, which is used by ScoreHMR (referred to as \textbf{HMR2.0b init.}), and 
(3) initializing with CameraHMR predictions, the state-of-the-art generalized model trained on the same dataset as our method (referred to as \textbf{CameraHMR init.}).

As shown in~\cref{tab:vsemdb}, our method outperforms ScoreHMR significantly when initialized from poses generated by the prior. 
When initializing with HMR2.0b poses, which often produce implausible bending poses (see~\cref{fig:ablation_point}), our PointDiT effectively corrects these erroneous poses and achieves a significantly larger improvement than ScoreHMR.
Furthermore, even when initialized with the SOTA pose regressor, CameraHMR, we observe a notable enhancement in the non-PA metrics, whereas ScoreHMR degrades the results.
This shows that incorporating user-specific shape information effectively reduces ambiguity in recovering body scale and orientation (\ie, in absolute 3D poses).

For completeness, we also supplement results on 3DPW (\cref{sec:res_3dpw}) and compare our performance to purely learning-based models (\cref{sec:res_learning}).
Our method either surpasses the state-of-the-art or ranks second-best, despite baseline methods often being trained on multiple datasets (\eg, NLF~\cite{sarandi2024nlf}). On the 3DPW dataset, we observe similar performance gains as on EMDB, albeit less pronounced.

\definecolor{Gray}{gray}{0.85}

\begin{table}[t]
    \centering
\resizebox{\columnwidth}{!}{%

\setlength\tabcolsep{4pt}
\begin{tabular}{l|cccc}
\toprule
Method               & MPJPE $\downarrow$ & MPJPE-PA $\downarrow$ & MVE $\downarrow$ & MVE-PA $\downarrow$ \\
\midrule
ScoreHMR~\cite{stathopoulos2024score} Sample init.  &  114.0 & 82.3 & 141.3 & 101.9  \\
\methodname~(Ours) Sample init. & 73.6 & 49.2 & 86.4 & 59.1  \\
\midrule
HMR2.0b~\cite{goel2023humans} init. & 117.2 & 77.9 & 140.2 & 93.9 \\
w/ SMPLify~\cite{Bogo:ECCV:2016}& - & 83.5\scriptsize{\red{(+5.6)}}  & - & - \\
w/ ScoreHMR~\cite{stathopoulos2024score}  & - & 76.5\scriptsize{\blue{(-1.4)}}  & - & - \\
w/ ScoreHMR*~\cite{stathopoulos2024score} & 105.5\scriptsize{\blue{(-11.7)}}  & 70.0\scriptsize{\blue{(-7.9)}} & 124.5\scriptsize{\blue{(-15.7)}} & 84.7\scriptsize{\blue{(-9.2)}}  \\ 
w/~\methodname~(Ours) & 73.2\scriptsize{\blue{(-44.0)}}  & 47.4\scriptsize{\blue{(-30.5)}} & 86.4\scriptsize{\blue{(-45.8)}} & 58.5\scriptsize{\blue{(-35.4)}}  \\
 \midrule
 CameraHMR~\cite{patel2025camerahmr} init. & 70.3 & 43.3 &  81.7 & - \\
 w/ ScoreHMR*~\cite{stathopoulos2024score} & 74.9\scriptsize{\red{(+4.6)}}  & 45.0\scriptsize{\red{(+1.7)}}  &  89.0\scriptsize{\red{(+7.3)}}  &  54.5 \\
 w/~\methodname~(Ours)  &\textbf{62.5}\scriptsize{\blue{(-7.8)}}&
 \textbf{42.4}\scriptsize{\blue{(-0.9)}} & \textbf{74.6}\scriptsize{\blue{(-7.1)}} &  \textbf{51.6}  \\

\bottomrule
\end{tabular}
}%
\caption{\textbf{ Pelvis-aligned local pose accuracy} on EMDB1. \emph{Top}: Body fitting using sampled pose from the body priors. \emph{Middle}: Body fitting using the HMR2.0b predictions as initialization, which is originally used in the ScoreHMR~\cite{stathopoulos2024score} paper. \emph{Bottom}: Body fitting using the CameraHMR predictions, which is the latest SOTA. * denotes re-run with the same shape and focal length.}

\label{tab:vsemdb}
\end{table}

\paragraph{Absolute pose accuracy.}
To fairly compare absolute pose accuracy in the camera coordinate system (\ie, without any alignment step), we provide all methods with the ground-truth camera focal length.
The results are summarized in~\cref{tab:cam_res}.
Interestingly, methods that achieve accurate pelvis-aligned poses (low MPJPE in~\cref{tab:vsemdb}) do not necessarily ensure absolute pose accuracy (high C-MPJPE/Pelvis Err.), \eg, CameraHMR.
We believe that this is due to the overfitting of pelvis-aligned objectives during training. 
While deterministic regressors can achieve lower errors in terms of local pose, the estimated shape and pelvis position might not always be plausible in 3D. C-MPJPE is a more suitable metric for evaluating the 3D pose accuracy in real-world scenarios, and our personalized fitting method has demonstrated improved performance on this metric.

\paragraph{Shape accuracy.}
To evaluate the accuracy of SHAPify, we use T-pose input images of 10 subjects from the EMDB dataset and compare our method with shapes extracted from SOTA methods. 
We report SHAPify's accuracy on EMDB in~\cref{tab:shape_res}. SHAPify consistently outperforms all baselines. 
Even on a simple T-pose image, existing methods predict shapes with significant errors, particularly when subjects wear loose-fitting clothing, such as jackets.
By incorporating user-specific measurements into the fitting process, we effectively address this ill-conditioned problem. Please find more results in Supp. Mat.~\cref{sec:res_shape}.
\begin{table}[t]
    \centering
\resizebox{\columnwidth}{!}{%
\begin{tabular}{l|cc}
\toprule

 Method         & Pelvis Err.(mm) $\downarrow$  & C-MPJPE(mm) $\downarrow$  \\
\midrule
 ScoreHMR~\cite{stathopoulos2024score} Sample init.   & 154.3 & 154.4 \\
 PHD (Ours) Sample init. & \textbf{91.5} & \underline{115.8} \\

 \midrule
 HMR2.0b~\cite{goel2023humans} init.    & 144.0 & 182.0 \\
 w/ ScoreHMR*~\cite{stathopoulos2024score}  & 180.6 \scriptsize{\red{(+36.6)}}  & 181.4 \scriptsize{\blue{(-0.6)}} \\
 w/ PHD (Ours)   & \underline{94.7} \scriptsize{\blue{(-49.3)}} & \textbf{112.6} \scriptsize{\blue{(-69.4)}} \\
 \midrule
 CameraHMR~\cite{patel2025camerahmr} init.  & 163.0 & 160.3 \\
 w/ ScoreHMR*~\cite{stathopoulos2024score}  & 154.3\scriptsize{\blue{(-5.7)}} & 154.4\scriptsize{\blue{(-5.9)}}\\
 w/ PHD (Ours)    & 130.9 \scriptsize{\blue{(-32.1)}} & 135.6  \scriptsize{\blue{(-27.4)}} \\
\bottomrule
\end{tabular}
}%
\caption{\textbf{Camera coordinates absolute pose accuracy} on EMDB1. Methods with a low pelvis-aligned error (see \cref{tab:vsemdb}) overfit local pose objectives and struggle with absolute pose.}

\label{tab:cam_res}

\end{table}

\begin{table}[t]
    \centering
    \footnotesize

\begin{tabular}{l|cc|cc}
\toprule
              &  \multicolumn{2}{c|}{Joint Error (mm)} &  \multicolumn{2}{c}{Vertex Error (mm)}     \\
  Method     &  Mean $\downarrow$ & Max $\downarrow$ &  Mean $\downarrow$ & Max $\downarrow$  \\
\midrule
Zero Shape & 28.41 & 52.72 & 29.84 &  60.18 \\
ScoreHMR~\cite{stathopoulos2024score} Mean &  29.07 & 55.37 &  29.32 & 60.23 \\
\midrule
 CameraHMR~\cite{patel2025camerahmr}    & 30.60 & 57.28 & 31.85 & 62.09 \\
 TokenHMR~\cite{dwivedi_cvpr2024_tokenhmr}    & 25.15 & 53.36 & 27.46 & 50.49 \\
 SHAPY~\cite{choutas2022accurate}        & 22.94 & 41.02 & 21.38 & 44.82 \\
  NLF~\cite{sarandi2024nlf}           & 19.36 & 36.37 & 20.61 & 41.46 \\
  SHAPify (Ours) w/o M.         & 13.97 & 29.30 & 14.25 & 33.44 \\
 SHAPify (Ours)     & \textbf{11.29} & \textbf{20.97} & \textbf{9.18} & \textbf{21.91} \\

\bottomrule
\end{tabular}

\caption{\textbf{Shape estimation comparison} on EMDB1. EMDB provides ground-truth SMPL shape from high-quality 3D scans.}

\label{tab:shape_res}
\end{table}

\subsection{Comparisons of Conditional 3D Body Prior}
\label{sec:compare_prior}
To verify the robustness of using PointDiT as the 3D body prior, we compare it with the pose prior in ScoreHMR.
The results are presented in~\cref{fig:big_fig} and~\cref{tab:prior_res}.
We first analyze the accuracy of image-conditioned body pose sampling. 
In~\cref{fig:big_fig} (left), ScoreHMR's prior fails to sample plausible body poses from input images of challenging poses, whereas our method produces plausible body poses that match the images. 
ScoreHMR requires relatively accurate pose initializations, as evident from our experiments and~\cref{fig:big_fig} (right).
When initializing with HMR2.0, which can produce implausible 3D poses, ScoreHMR struggles to recover from it due to the ambiguity of 2D keypoint reprojection. 
In contrast, PointDiT provides effective 3D guidance during fitting, helping to correct implausible 3D poses.

\subsection{Ablation Studies}
\label{sec:ablation}
\begin{figure}[t]
\centering
\includegraphics[width=\linewidth]{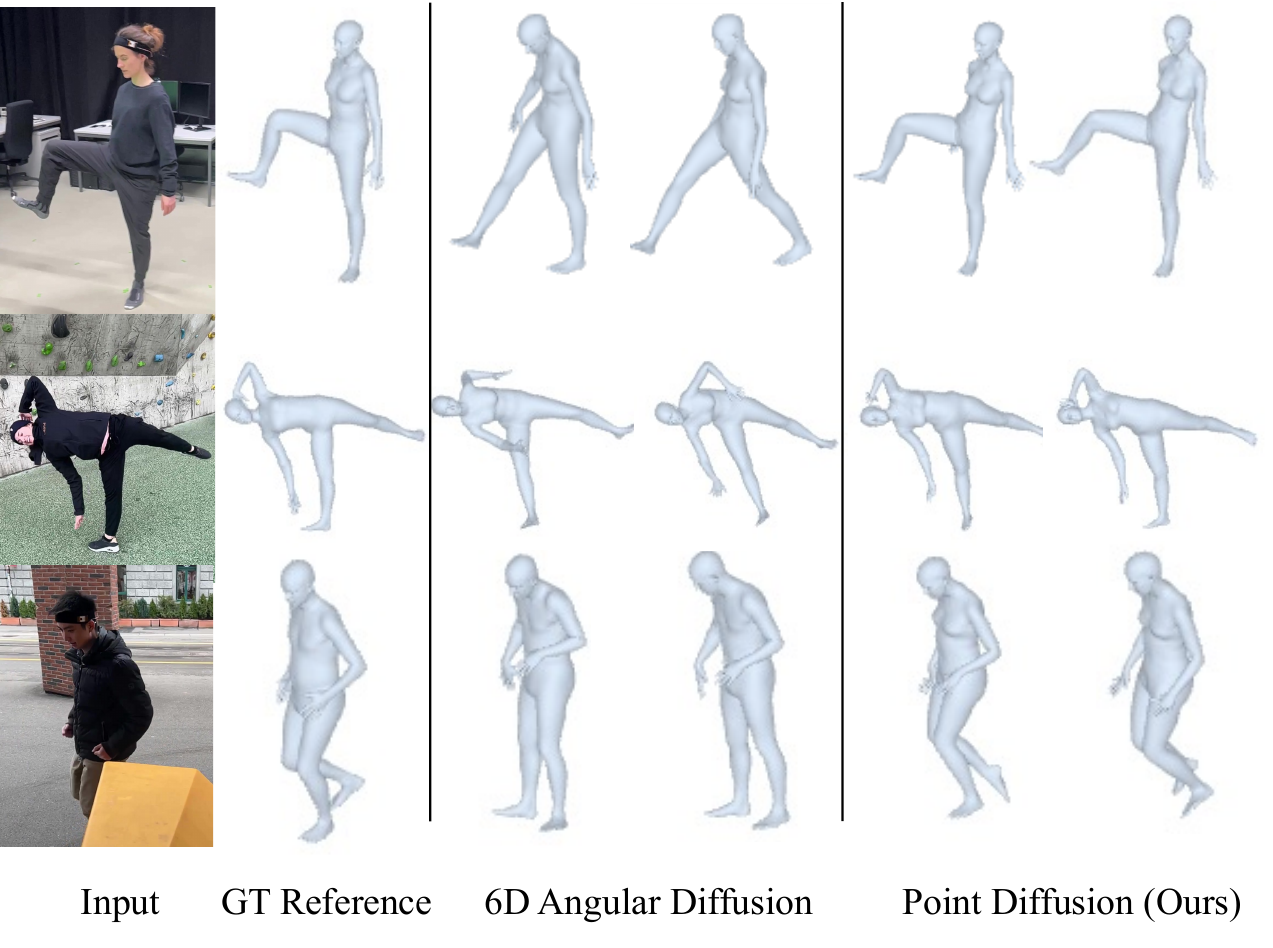}
\caption{\textbf{Comparison of pose representations} used by the prior. We show two random samples each. An angular-based representation produces less desirable poses.}
\label{fig:compare_prior}
\end{figure}

\paragraph{Point cloud vs. joint rotation representation.}
To evaluate the benefits of using point clouds as the 3D body representation over joint rotations, we trained a variant of the PointDiT model. This variant maintains the same input conditions and network architecture but outputs 6D joint rotations~\cite{Cho_2023_ICCVW} (referred to as \textbf{6D Angular}).
The results presented in~\cref{fig:compare_prior} and~\cref{tab:prior_res} indicate that under identical image conditions, the 6D joint rotations result in greater errors in sampled poses, particularly for uncommon poses. We attribute this to the weak correlation between 2D image features and joint rotations.
ScoreHMR addresses this issue by utilizing features extracted from a deep layer of a learned 3D pose regressor~\cite{Kocabas_PARE_2021}.
While these features are more closely related to joint rotations, they still suffer from the regressor's limitations of being less effective on uncommon poses.
For example, in~\cref{fig:big_fig} (left), ScoreHMR cannot sample a reasonable pose for images of out-of-distribution body poses.
In contrast, by using 2D image features with the point cloud representation, our model can handle uncommon poses more effectively (see~\cref{fig:big_fig} Ours Sampling).

\begin{table}[t]
    \centering
    \footnotesize

\begin{tabular}{lccc}
\toprule
 Method & MPJPE & PA-MPJPE & PA-MVE \\
\midrule
 6D Angular~\cite{Cho_2023_ICCVW}  & 177.9 & 125.2  & 154.8 \\
 ScoreHMR~\cite{stathopoulos2024score}  & 150.0 & 102.3 & 128.0 \\
 Points (Ours)   & \underline{75.6} & \underline{52.1} & \underline{62.1} \\
 \midrule
 ScoreHMR~\cite{stathopoulos2024score} Sample init. & 114.0 & 82.3 & 101.9 \\
 PHD (Ours) Sample init.    & \textbf{73.6} & \textbf{49.2} & \textbf{59.1}  \\

\bottomrule
\end{tabular}

\caption{\textbf{Comparison of 3D body priors} on EMDB1. \textit{Top:} Performance of different pose representations. \textit{Bottom:} Performance when fitting the sampled initializations. We note that simple sampling from our prior already outperforms ScoreHMR's fitting.}

\label{tab:prior_res}

\end{table}

\paragraph{Effectiveness of personalization.}
We analyze how body shape affects the accuracy of pose fitting. In~\cref{tab:ablation}, we evaluate four different shape configurations: zero shape, per-subject mean shape (from ScoreHMR), fitted shape from SHAPify (w/ and w/o measurements), and the ground-truth shape.
Surprisingly, using the mean shape results in poorer fitting performance compared to using the zero shape.
Note that averaging the shape is a common practice~\cite{guo2023vid2avatar,shin2023wham,stathopoulos2024score}. 
Moreover, more accurate body shapes not only improve pelvis-aligned poses but also enhance absolute pose accuracy.
We visualize this effect in Supp. Mat.~\cref{sec:res_pelvis}.

\begin{figure}[t]
\centering
\includegraphics[width=\linewidth]{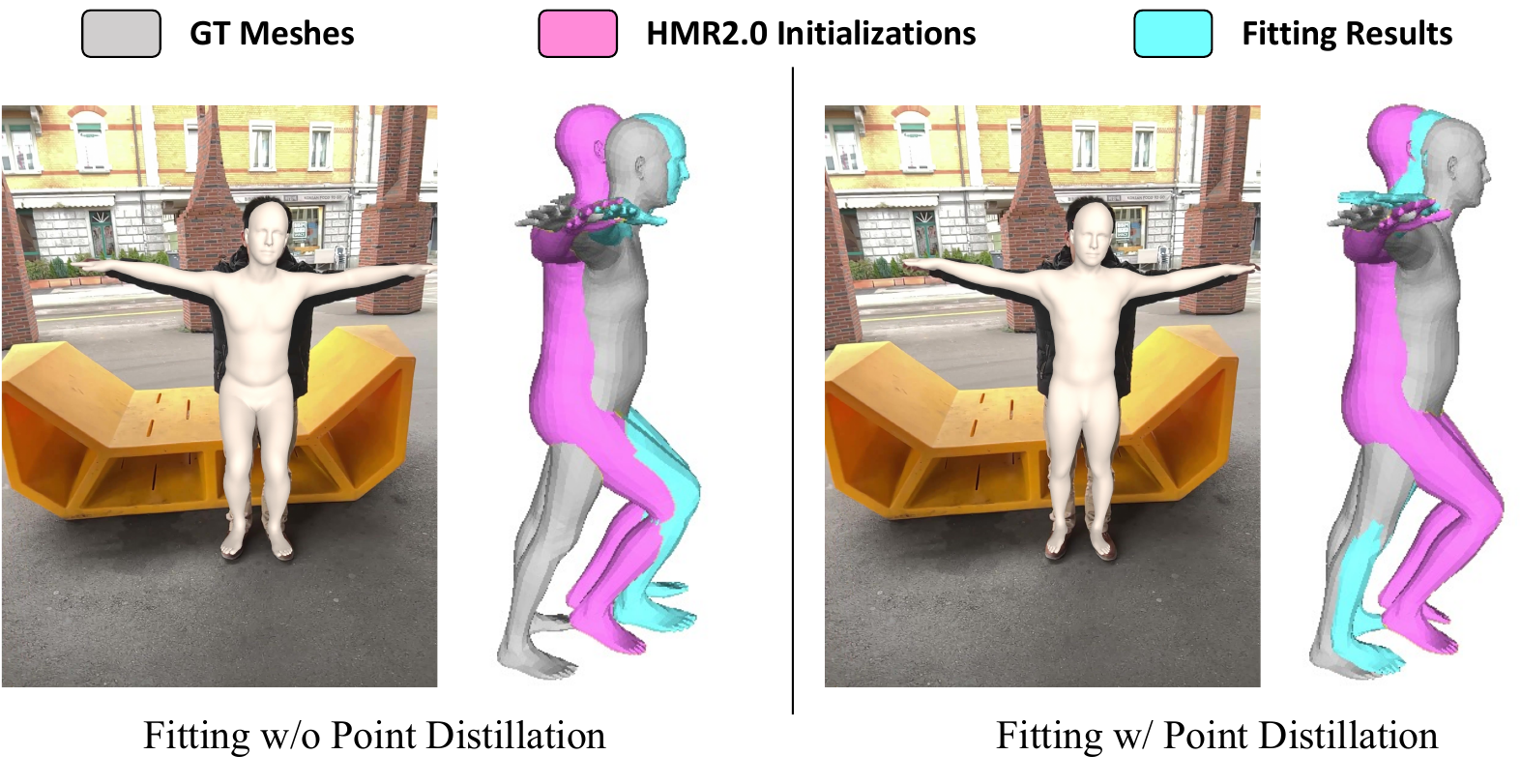}
\caption{\textbf{Point distillation} successfully corrects bad input poses that result from over-optimization to 2D keypoints.}
\label{fig:ablation_point}
\end{figure}

\paragraph{Effectiveness of point distillation.}
Finally, we demonstrate the efficacy of point distillation sampling in body fitting.
In~\cref{fig:ablation_point}, we highlight a common issue of knee bending observed in many existing methods (here HMR2.0b).
Such poses may exhibit minimal 2D keypoint projection errors but are highly implausible in 3D.
Relying solely on 2D visual cues for pose refinement is insufficient to resolve this issue.
As shown in~\cref{fig:ablation_point} (right), incorporating point distillation sampling rectifies this problem, ensuring both 3D plausibility and 2D alignment. 
Consequently, the accuracy of local poses is largely improved (see~\cref{tab:ablation} (bottom)).
\begin{table}[t]
    \centering
    \scriptsize

\setlength\tabcolsep{1.8pt} 
\begin{tabularx}{\hsize}{lccccc}
\toprule
 Method            & Shape Err. J/V &  MPJPE$\downarrow$  & MVE$\downarrow$  &  C-MPJPE &  Pelvis Err.$\downarrow$ \\
\midrule
 Mean Shape & 29.1 / 29.3 & 81.0  & 94.6 & 177.3 & 163.6\\
 Zero Shape & 28.4 / 29.8 & 79.8 &  93.2  &  170.8 & 156.8\\
 SHAPify w/ M. & 11.3 / 9.2 & 73.6  & 86.4  & 115.8 & 91.5\\
 SHAPify w/o M. & 13.9 / 14.2 & 73.2  & 86.0  & 116.3 & 95.0 \\
 GT Shape & -  & 72.5 & 85.2 & 110.4 & 84.7\\
 \midrule
 w/o Point Distillation & - & 75.8 & 91.7& 111.4 & 83.3 \\
 w/ Point Distillation & -  & 72.5 & 85.2  & 110.4 & 84.7\\
\bottomrule
\end{tabularx}

\caption{\textbf{Ablation studies} on EMDB1.~Top: Effectiveness of different body shapes. Bottom: Importance of our point distillation.}
\label{tab:ablation}
\end{table}

\section{Conclusion}
\label{sec:discuss}
\methodname~represents a step forward from generic models towards personalized 3D human body recovery. By decoupling the traditional regression pipeline into a shape calibration and a body fitting procedure, PHD effectively leverages user-specific identity information to achieve more accurate and robust 3D pose estimates. The core of our method is a powerful shape-conditioned 3D prior, implemented as a point-based diffusion transformer, which guides the body fitting process. Our results showcase that~\methodname~is not only highly versatile but also holds potential to drive future human-centric perceptual AI systems.
\vspace{-.5em}
\paragraph{Acknowledgements.} This work was partially supported by the Swiss SERI Consolidation Grant "AI-PERCEIVE".

\clearpage


{
    \small
    \bibliographystyle{ieeenat_fullname}
    \bibliography{main}
}
\maketitlesupplementary

\section{Implementation Details}
\begin{figure}[t]
\centering
\includegraphics[width=\linewidth]{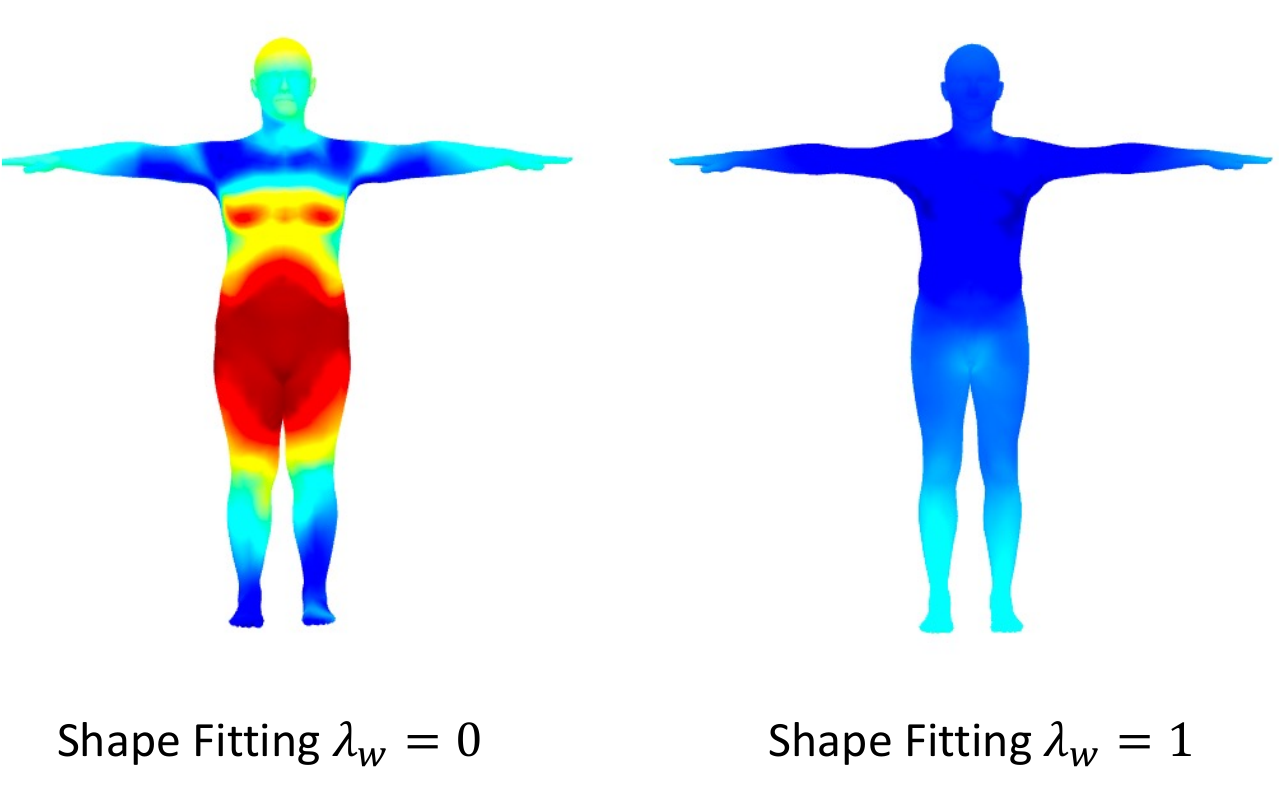}
\caption{Importance of weight regularization. While the body height is correct, without weight regularization, the fitting is prone to converge at high BMI shape parameters.}
\label{fig:belly}
\end{figure}

\begin{figure}[t]
\centering
\includegraphics[width=\columnwidth]{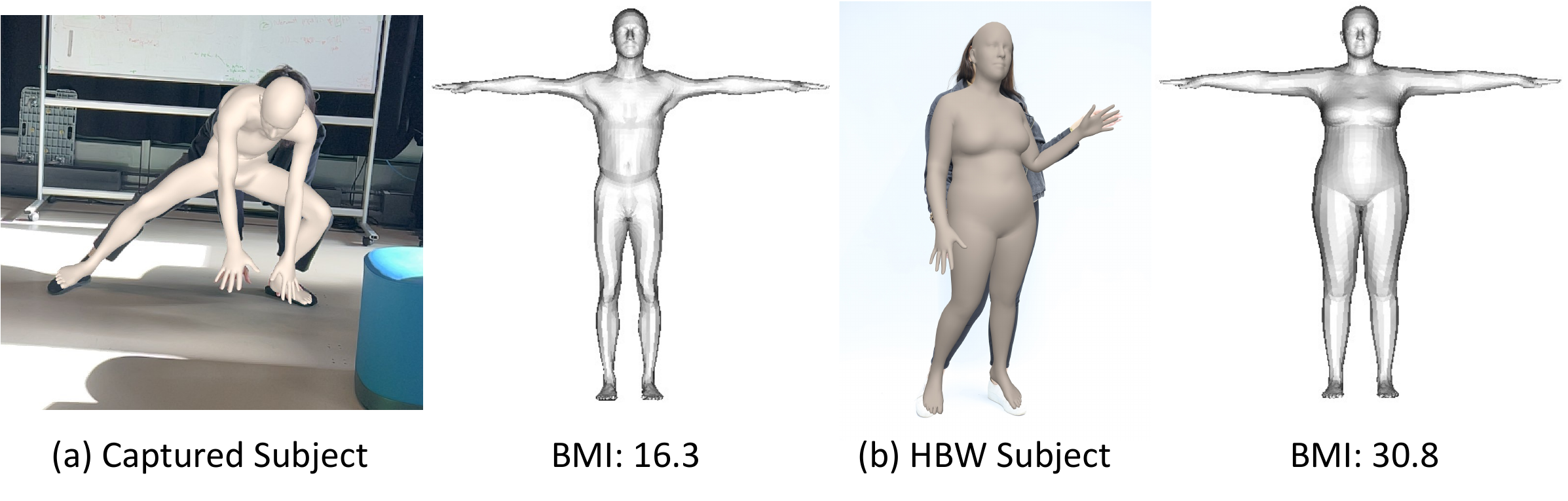}
\caption{SHAPify is also capable of estimating the shapes of subjects with high and low BMI values, which allows us to perform customized pose estimation for diverse human subjects. }
\label{fig:extreme}
\end{figure}

\subsection{SHAPify Details}
\label{sec:shapify_details}
In the optimization of SHAPify, we initialize the pose parameters $\theta$ as the rest pose (T-/I-pose) and the pelvis position $\mathbf{p}$ to:
\begin{equation}
    \mathbf{p} = [\frac{x_p-c_x}{f}Z, \frac{y_p-c_y}{f}Z, Z],
\end{equation}
where $f$ is the camera focal length and $(x_p,y_p), (c_x,c_y)$ are the pelvis pixels and camera center on the image, respectively. $Z$ is the depth of the pelvis, which we approximate as $Z = f \cdot SW_\text{SMPL} / SW_\text{kp}$. Note that $SW_\text{SMPL}$ is the shoulder width of the SMPL mean shape, and $SW_\text{kp}$ is the length of shoulder keypoints on the 2D image. This approximation holds because the horizontal line on the image is not affected by the pitch angles of global orientation, and we can assume that the roll and yaw angles of the pelvis orientation $\phi$ are typically small in the frame of rest poses. We also initialize the roll and yaw angles to $0$ and update them with small learning rates.

In the regularization term of SHAPify, we calculate body heights and weights from the SMPL body meshes. The heights ($H(\cdot)$) are calculated as the distance between the top of the head (Vertex$\#$ 411) to the center of feet (Vertex$\#$ 3439, 6839). The heights ($W(\cdot)$) are the volume of human body meshes multiplied by the body density (985 $kg/m^3$). We use $\lambda_\beta = 0.1$, $\lambda_h = 100$, and $\lambda_w = 10$ if the body measurements are available, and $\lambda_\beta = 1$, $\lambda_h = 1$, and $\lambda_w = 1$ if not. Without using the body measurements, our method achieved a 14mm joint error and a 13mm vertex error (\cref{tab:shape_res}), which is only slightly higher than using the body measurements. Moreover, we found the body weight regularization term crucial. Without such a constraint, the fitting is prone to converge to shapes with large bellies. (See~\cref{fig:belly})

\begin{figure}[t]
\centering
\includegraphics[width=\linewidth]{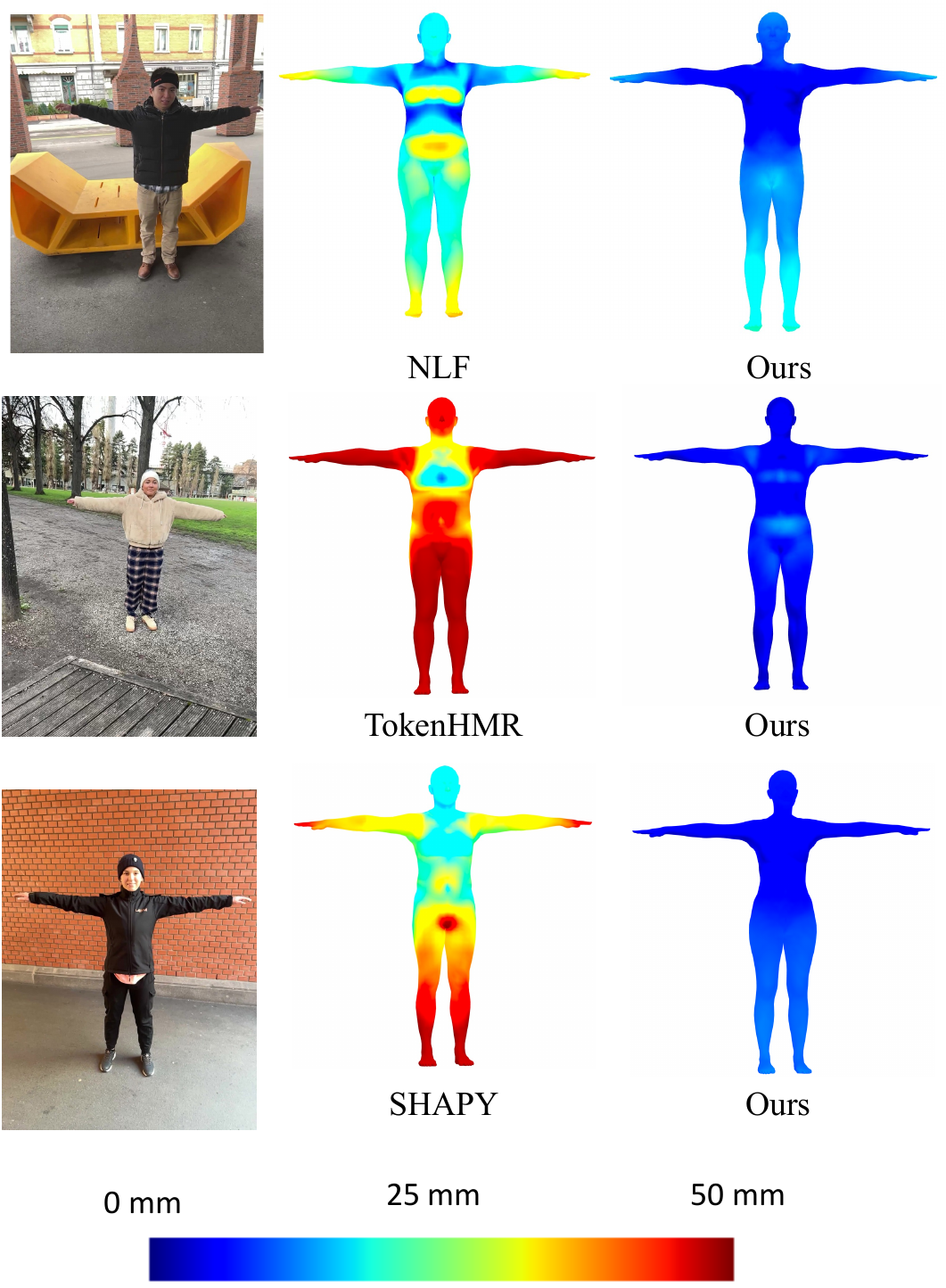}
\caption{\textbf{Visualization of shape estimation errors} on sequences in EMDB. See also~\cref{tab:shape_res} for quantitative results.}
\label{fig:compare_shape}
\end{figure}

\subsection{PointDiT Architecture and Training}
\label{sec:pointdit_details}
\begin{figure}[t]
\centering
\includegraphics[width=\linewidth]{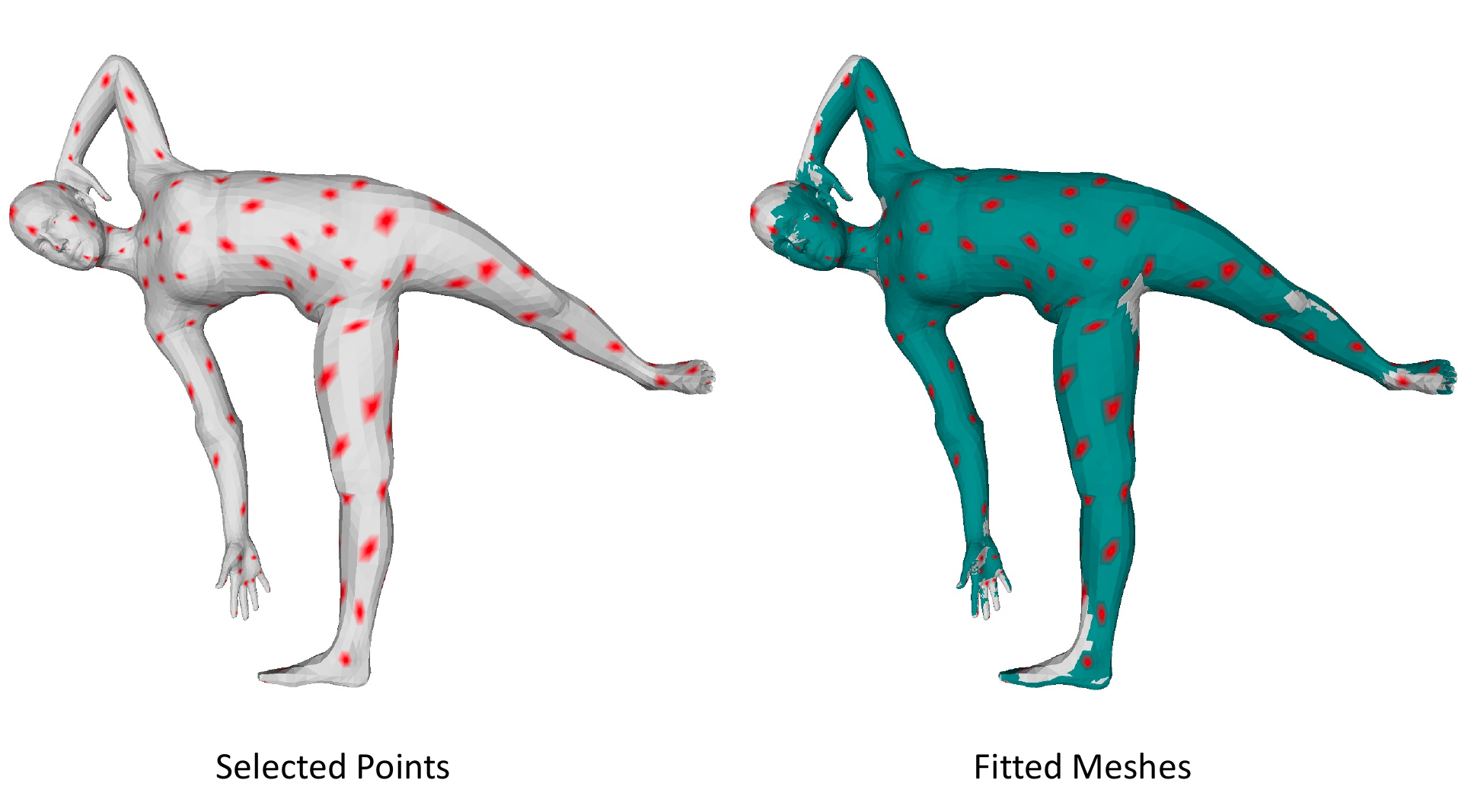}
\caption{Visualization of selected body surface points. \emph{Left}: We use the 238 vertices and 45 joints to fit the SMPL parameters. \emph{Right}: The fitted body mesh (in green color) is highly aligned with the original ground-truth body mesh.}
\label{fig:fitter}
\end{figure}

\begin{figure}[t]
\centering
\includegraphics[width=\linewidth]{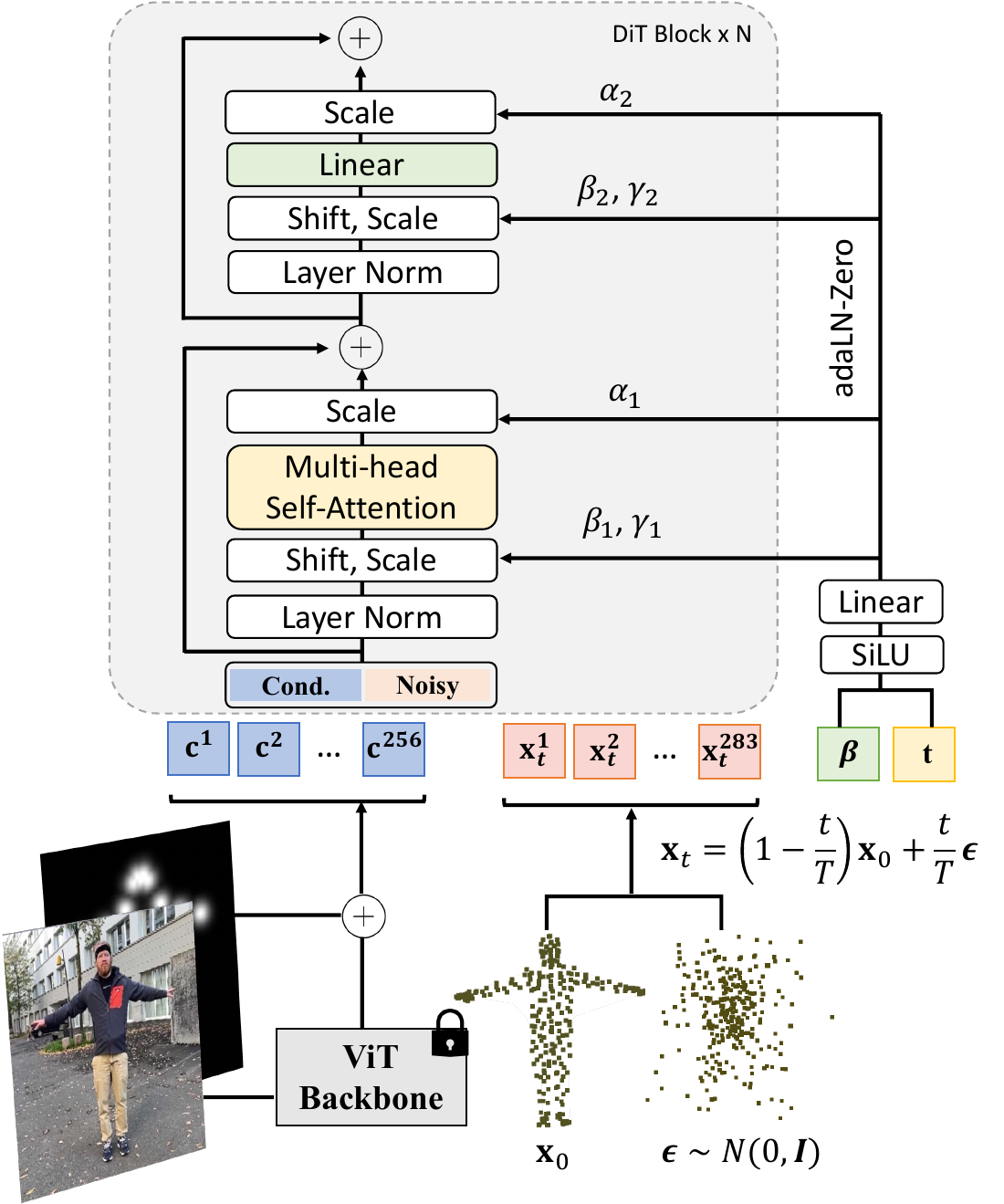}
\caption{\textbf{Network architecture of PointDiT}. Our point clouds are made up of 45 SMPL joints and 238 vertices from the SMPL surface. This was chosen in accordance with the accuracy of the SMPL Fitter in NLF \cite{sarandi2024nlf}. Hence, the point clouds are of 283 dimensions. We use the rectified flow formulation \cite{liu2022flow,sd3} in the diffusion model to reduce the number of denoising steps as described in the main paper.}
\label{fig:pointdit}
\end{figure}

We show the detailed network architecture of PointDit in~\cref{fig:pointdit}. More specifically, the PointDiT model contains 20 DiT blocks and is operated at a dimension of 512. We use the frozen ViTBackbone~\cite{xu2022vitpose} to extract $256\times256$ image features of the shape of $16\times16\times1280$ and heatmaps of the shape of $64\times64\times17$. We add the image features, heatmaps, and positional embeddings together to obtain the final conditional features $\textbf{c}$. Our body point clouds are made up of 45 SMPL joints and 238 vertices from the SMPL surface (see~\cref{fig:fitter}). This was chosen by the accuracy of the Point Fitter in NLF~\cite{sarandi2024nlf}. The point clouds used for diffusion are of the shape $283\times3$, and we normalize the points to zero mean and unit variance before adding noise. We use the rectified flow formulation~\cite{liu2022flow,sd3} in the diffusion model to reduce the number of denoising steps during inference, as described in the main paper. Both conditional features and point clouds are projected to 512-dimensional tokens and fed into the transformer. In the output layer, we project the tokens back to 3-dimensional points and de-normalize them. The image conditional tokens are only used for self-attention conditioning and are discarded in the final layer. 

We train our PointDiT model using the synthetic BEDLAM dataset. We apply standard data augmentations~\cite{Black_CVPR_2023, dwivedi_cvpr2024_tokenhmr} to the provided image crops and ground-truth annotations. Our training consists of two stages. In the first stage, we set all conditional shape parameters $\beta$ to zero, allowing the model to focus on sampling body point clouds corresponding to the conditional images. In the second stage, we learn the correct body shape of point clouds using the ground-truth shape parameters as conditions. For training, we utilize a batch size of 512 images and set the learning rate to $10 ^{-5}$ with the AdamW optimizer. The training takes approximately 1 day for the first stage, with 12K iterations, and another 2 days for the second stage, with 30K iterations, on 8 NVIDIA V100 GPUs. The training scheduler and reweighting factors follow the same configuration as in Stable Diffusion 3~\cite{sd3}. We employ a dropout rate of 0.05 for the image and shape conditioning.

\subsection{Point Distillation and Body Fitting Details}

In the body fitting stage, we can initialize pose $\theta_0$ and global orientation $\phi_0$ with either our sampled points $\mathbf{x}_0$ or the results from a regressor. To initialize the pelvis position $\mathbf{p}_0$, we solve the weighted least squares problem by plugging in  $\theta_0, \phi_0$ to~\cref{eq:fitting_kp}. Afterward, we optimize 100 iterations per image with a learning rate of $3e-3$ for optimizing $\theta$ and $1e-3$ for optimizing $\phi, \mathbf{p}$ with the AdamW optimizer. The $\lambda_\text{data}$ is set to 1.0 and the $\lambda_\text{prior}$ is set to 100, ($\lambda_{p}$,  $\lambda_{\phi}$,  $\lambda_{\theta}$) are set to (0.1, 0.1, 1.0). Every 10 iterations, we resample the points again from the fitted meshes. 

\begin{figure*}[ht]
\centering
\includegraphics[width=\linewidth]{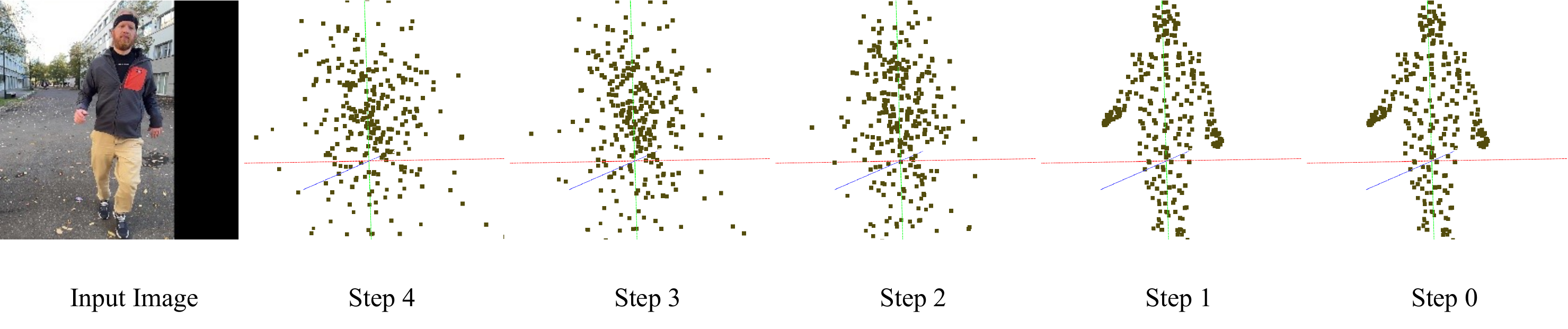}
\caption{\textbf{Example of the body points at each denoising time step.} Using the rectified flow formulation allows us to efficiently sample body point clouds in as few as 5 denoising steps. }
\label{fig:point}
\end{figure*}

\subsection{Inference}
We extract 2D keypoints using Sapiens~\cite{khirodkar2024sapiens}, and we pre-process keypoints and bounding boxes (image crops) before inference.
SHAPify is a lightweight optimization algorithm that can be run on a CPU and takes roughly 1 second for each image. For body fitting inference, we tested our model on a single NVIDIA RTX 3090 GPU. We set the number of denoising steps to $T=5$, and it takes approximately 1 second per frame without any parallelization (batch size $=1$).
As a reference, ScoreHMR~\cite{stathopoulos2024score} requires $T=20$ denoising steps and takes about 3 seconds per frame under the same data setup. 
\section{More Experimental Results}
\label{sec:supp_exp}
\begin{table}[t]
    \centering
    \small
\setlength\tabcolsep{4pt}
\begin{tabular}{lcccc}
\toprule
 Method & 
          \begin{tabular}[c]{@{}c@{}} MPJPE \\$\downarrow$ (mm) \end{tabular}  &
          \begin{tabular}[c]{@{}c@{}} PA-MPJPE \\$\downarrow$ (mm) \end{tabular}  &
          \begin{tabular}[c]{@{}c@{}} MVE  \\$\downarrow$ (mm) \end{tabular} & 
          \begin{tabular}[c]{@{}c@{}} PA-MVE  \\ $\downarrow$(mm) \end{tabular} \\
\midrule
 IK (24 joints)~\cite{zuo2021sparsefusion}   & 67.8 & 49.3  & 81.9 & 60.3 \\
 \midrule
 25\% Points  & 63.8 & 45.3 & {78.1} & {59.0}  \\
 50\% Points  & 63.2 & 44.7 & 73.7 & 53.6  \\
 100\% Points & \textbf{62.6} & \textbf{44.4} & \textbf{72.9} & \textbf{53.1}  \\
\bottomrule
\end{tabular}
\caption{\textbf{Effectiveness of Point Fitter.} We report the fitting accuracy on the EMDB ``P1-14" sequence using IK and Point Fitter. We also analyze the robustness of Point Fitter by dropping out surface points during the fitting process.}
\label{tab:ablation_pts}
\end{table}

\subsection{Shape Estimation}
\label{sec:res_shape}
We visualize the results of SHAPify against existing methods on EMDB in~\cref{fig:compare_shape}. Even on a simple T-pose image, existing methods predict shapes with significant errors, particularly when subjects wear loose-fitting clothing, such as jackets.

EMDB mainly contains subjects with moderate body shape. Since real-world data of extreme body shapes is limited, in~\cref{fig:extreme}, we visualize the SHAPify and PHD's fitting results on a newly recorded slim subject and another high-BMI subject from the HBW~\cite{choutas2022accurate} dataset. Because PointDiT is trained on BEDLAM, which contains body shapes across a wide BMI range (17.5 to 42.5), our method is able to generalize to human subjects with diverse body shapes.

\subsection{Effectiveness of Point Fitter}
In~\cref{tab:ablation_pts}, we analyze the effect of the point cloud size by retaining 25\% and 50\% of the points for fitting. Furthermore, Inverse Kinematics (IK)~\cite{zuo2021sparsefusion, ludwig2024leveraging} is commonly used for fitting 3D joints back to the parameter space. We also design a joint-only baseline by replacing the Point Fitter with an open-source IK solver~\cite{zuo2021sparsefusion}. The joint-only IK is less accurate than Point Fitter and incurs expensive optimization loops that slow down the method (15 seconds per frame). In contrast, the Point Fitter is faster but requires more points to ensure accuracy. When only 25\% of points are retained, we observe a clear increase in MVE. Overall, our design is both efficient and accurate.

\definecolor{Gray}{gray}{0.85}

\begin{table}[t]
    \centering
\resizebox{\columnwidth}{!}{%

\setlength\tabcolsep{4pt}
\begin{tabular}{l|cccc}
\toprule
Method               & MPJPE $\downarrow$ & MPJPE-PA $\downarrow$ & MVE $\downarrow$ & MVE-PA $\downarrow$ \\
\midrule
ScoreHMR~\cite{stathopoulos2024score} Sample init.  &  94.3 & 66.6  & 122.3 & 94.7  \\
\methodname~(Ours) Sample init. & 84.2 &  51.0 & 100.4 &  67.6  \\
\midrule
HMR2.0b~\cite{goel2023humans} init. &  81.7 & 54.2 & 93.5& 67.7 \\
w/ SMPLify~\cite{Bogo:ECCV:2016}& - & 60.1\footnotesize{\red{(+5.9)}} & - & - \\
w/ ScoreHMR~\cite{stathopoulos2024score}  & - & 51.1\footnotesize{\blue{(-3.1)}} & - & - \\
w/ ScoreHMR*~\cite{stathopoulos2024score} & 75.7\footnotesize{\blue{(-6.0)}} & 51.2\footnotesize{\blue{(-3.0)}}  & 87.5\footnotesize{\blue{(-6.0)}} &  65.4\footnotesize{\blue{(-2.3)}}   \\ 
w/~\methodname~(Ours) & 80.5\footnotesize{\blue{(-1.2)}} & 44.4\footnotesize{\blue{(-9.8)}} & 93.3\footnotesize{\blue{(-0.2)}} & 57.3\footnotesize{\blue{(-10.4)}}  \\
 \midrule
 CameraHMR~\cite{patel2025camerahmr} init. & 62.1 & 38.5 & 72.9 & -  \\
 w/ ScoreHMR*~\cite{stathopoulos2024score} & 59.6\footnotesize{\blue{(-2.5)}}  & 38.0\footnotesize{\blue{(-0.5)}}  & 71.5\footnotesize{\blue{(-1.4)}} & 51.1  \\
 w/~\methodname~(Ours)  &  \textbf{59.4}\footnotesize{\blue{(-2.7)}} &
 \textbf{37.5}\footnotesize{\blue{(-1.0)}} &
 \textbf{71.3}\footnotesize{\blue{(-1.6)}} &
 \textbf{50.9} \\

\bottomrule
\end{tabular}
}%
\caption{\textbf{ Pelvis-aligned local pose accuracy} on 3DPW (14 joints). \emph{Top}: Body fitting using sampled pose from the body priors. \emph{Middle}: Body fitting using the HMR2.0b predictions as initialization, which is originally used in the ScoreHMR~\cite{stathopoulos2024score} paper. \emph{Bottom}: Body fitting using the CameraHMR initialization. * denotes re-run with the same input shape and focal length.}

\label{tab:vs3dpw}
\end{table}

\begin{table*}[t]
    \centering
    \small

\resizebox{1.0\textwidth}{!}{
\begin{tabular}{l|cccc|cccc}
\toprule
                     & \multicolumn{4}{c|}{EMDB~\cite{kaufmann2023emdb}}         & \multicolumn{4}{c}{3DPW~\cite{von2018recovering}}          \\
\midrule
Method               & MPJPE $\downarrow$ & MPJPE-PA $\downarrow$ & MVE $\downarrow$ & MVE-PA $\downarrow$  & MPJPE $\downarrow$ & MPJPE-PA $\downarrow$ & MVE $\downarrow$ & MVE-PA $\downarrow$ \\

\midrule
HMR2.0b~\cite{goel2023humans} &  117.4 & 78.0 & 140.5 & 94.0 & 81.8 & 54.4 & 93.5 & 67.8 \\
PARE~\cite{Kocabas_PARE_2021} & 113.9 & 72.2 & 133.2 & 85.4  & 74.5 & 46.5 & 88.6 & - \\
HMR2.0a~\cite{goel2023humans} &  98.3 & 60.7 &  120.8 & - &   69.8 & 44.4 & 82.2 & - \\
TokenHMR~\cite{dwivedi_cvpr2024_tokenhmr} &  88.1 & 49.8 &  104.2 & - &  70.5 & 43.8 & 86.0 & - \\
CameraHMR~\cite{patel2025camerahmr} &  70.3 & 43.3 &  81.7 & - &  62.1 & 38.5 & 72.9 & - \\
NLF~\cite{sarandi2024nlf} & \underline{68.4} & \textbf{40.9} & \underline{80.6} & \textbf{51.1} & \textbf{59.0} &  \textbf{36.5} & \textbf{69.7} & \textbf{48.8} \\
  
\midrule
 LGD~\cite{song2020humanbodymodelfitting}   & 115.8   & 81.1  & 140.6 & 95.7 & - & 59.8 & - & - \\
 ReFit~\cite{wang23refit} & 88.0   & 58.6  & 104.5 & - & 65.3 & 40.5  & 75.1 & - \\
 WHAM*~\cite{shin2023wham} & 79.7   & 50.4  & 94.4 & - & \textbf{57.8}* & \textbf{35.9}*  & \textbf{68.7}* & - \\
 ScoreHMR~\cite{stathopoulos2024score} (CameraHMR init.) & 74.9 & 45.0 &  89.0 & 54.5 & 59.6 & 38.0 & 71.5 & 51.1 \\
\midrule
Ours (Sample init.) & 73.6 & 49.2 & 86.4 & 59.1 & 84.2 &  51.0 & 100.4 &  67.6  \\
Ours (HMR2.0b init.) & 73.2 & 47.4 & 86.4 & 58.5 & 80.5 & 44.4 & 93.3 & 57.3  \\
Ours (CameraHMR init.)  &\textbf{62.5} & \underline{42.4} & \textbf{74.6} &  \underline{51.6} & \underline{59.4} & \underline{37.5} & \underline{71.3} & \underline{50.9} \\
 
\bottomrule
\end{tabular}
}

\caption{\textbf{Comparisons of body fitting with learning-based methods on in-the-wild benchmarks}. * indicates fine-tuning on 3DPW, \textbf{bold} is best result, \textit{underlined} second best. Our method, initialized with CameraHMR, either beats the state of the art or performs second best with a narrow margin. This is remarkable as NLF was trained on multiple datasets, while we only trained on synthetic BEDLAM.}

\label{tab:pose}
\end{table*}

\begin{figure*}[t]
\centering
\includegraphics[width=\linewidth]{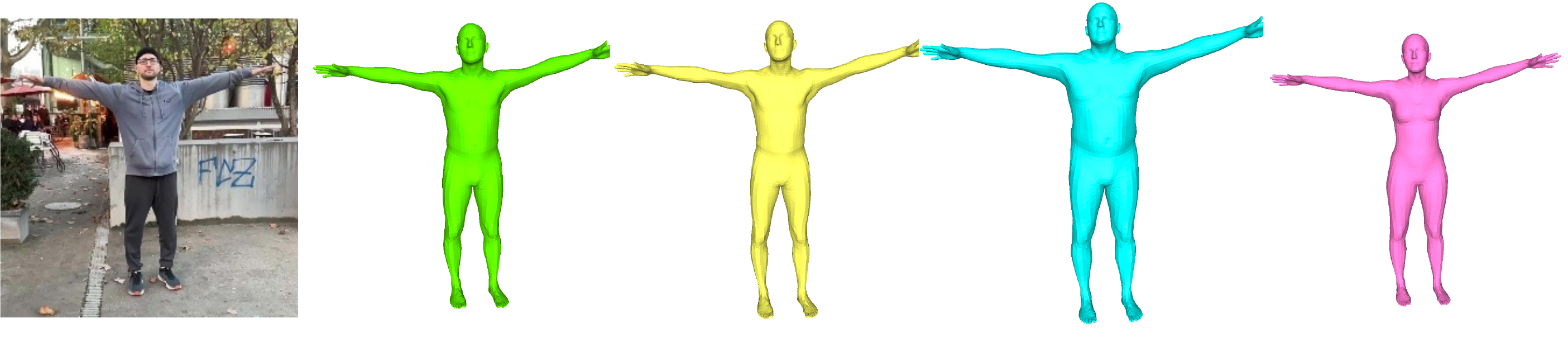}
\caption{\textbf{Example of shape conditioning.} Give the same input image but different $\beta$ for conditioning, PointDiT can sample point clouds in different body shapes. We utilize the Point Fitter to convert point clouds into body meshes for visualization. 
}
\label{fig:betas}
\end{figure*}

\subsection{Results on 3DPW}
\label{sec:res_3dpw}

We conduct similar evaluation in~\cref{sec:compare_all} on the 3DPW dataset~\cite{von2018recovering}. Note that for this evaluation benchmark, we don't have access to the body measurements of the subjects, and we use the estimated shapes for evaluation. Overall, we observed similar performance improvements as on EMDB, but less pronounced. We identified two major issues with this benchmark.

First, as noted in EMDB~\cite{kaufmann2023emdb}, performance on 3DPW has saturated due to its limited pose diversity, which motivates the need for EMDB as a diverse and challenging benchmark. Many recent methods report trends consistent with ours, \ie, more minor improvements on 3DPW compared to EMDB. Second, unlike EMDB, which uses EM sensors and SLAM to reconstruct accurate 3D camera trajectories and body poses, 3DPW estimates them through a joint optimization process based on 2D keypoints and IMU sensors. This approach introduces systematic errors of approximately 26 mm, as reported in the original paper, resulting in a biased pose distribution. Since this distribution deviates from that of synthetic training data, the learned prior becomes less effective during the fitting stage. Consequently, recent methods~\cite{sarandi2024nlf, shin2023wham} often finetune on the 3DPW training set to better align with its data distribution. Overall, we believe that consistent performance gains on EMDB without finetuning are more reliable indicators of robustness to diverse and challenging poses.

\subsection{Comparison to Learning-Based Methods}
\label{sec:res_learning}

We extensively compare our method against learning-based approaches in~\cref{tab:pose}. Notably, our model is trained exclusively on the synthetic BEDLAM dataset (1M images), whereas other methods are trained on a combination of real-world datasets such as Human3.6M~\cite{h36m_pami}, MPI-INF-3DHP~\cite{mono-3dhp2017}, and in the case of NLF~\cite{sarandi2024nlf}, over 40 datasets.

On the EMDB benchmark, when initializing our optimization with poses randomly sampled from PointDiT, we observe slightly higher errors compared to CameraHMR. This is likely due to domain gaps between the synthetic training images and real test-time images. However, when using CameraHMR for pose initialization, our method either outperforms the state-of-the-art (NLF~\cite{sarandi2024nlf}) or comes in a close second. On the 3DPW dataset, initializing with PointDiT-sampled poses yields suboptimal performance, primarily due to the biased pose distribution discussed in~\cref{sec:res_3dpw} and the domain mismatch between the training and test images. To address this, we utilize CameraHMR for pose initialization, achieving the second-best performance on 3DPW. It is worth noting that the (PA-)MPJPE values on 3DPW have saturated and are close to the reported systematic error (26mm). As such, it is difficult to determine whether improvements reflect actual accuracy gains or overfitting to inherent dataset biases. Consequently, we argue that the evaluations on EMDB provide a more meaningful reflection of true pelvis-aligned 3D pose accuracy.

\subsection{DiTPose Sampling}
In~\cref{fig:point}, we visualize the point clouds denoising process of our PointDiT model. Our model leverages the rectified flow formulation to train a diffusion model, enabling sampling of body point clouds in as few as 5 denoising steps. In~\cref{fig:betas}, we demonstrate the effectiveness of shape conditioning in our PointDiT model. Given the same input images but with different conditional shape parameters, our model generates diverse body shapes that correspond to the body poses described in the input images. 

\begin{figure*}[t]
\centering
\includegraphics[width=\linewidth]{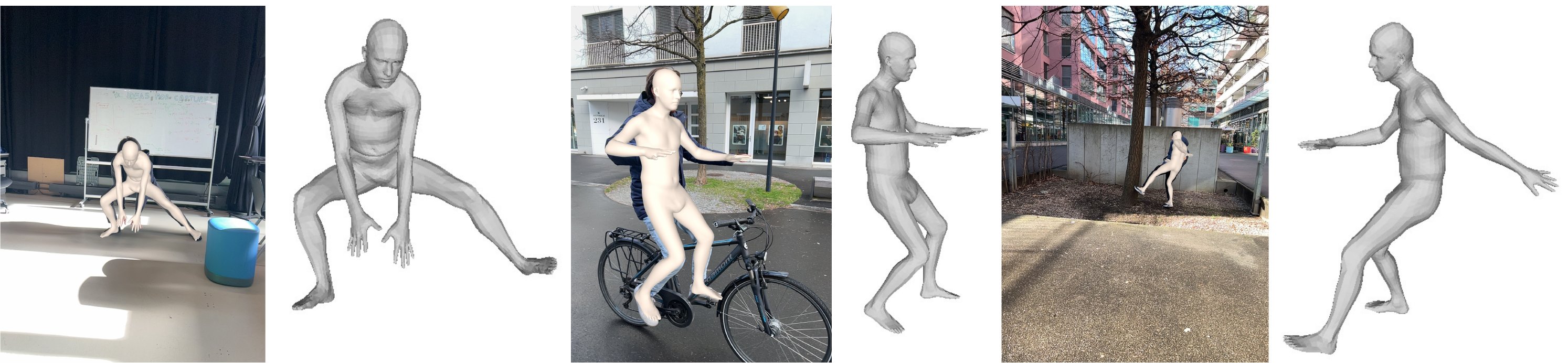}
\caption{\textbf{Applications on in-the-wild capturing}. Given the body measurements of human subjects, our method enables in-the-wild human performance capture using a modern smartphone.}
\label{fig:wild}
\end{figure*}

\subsection{Effect of Shape on Pelvis Position}
\label{sec:res_pelvis}
\begin{figure}[t]
\centering
\includegraphics[width=\linewidth]{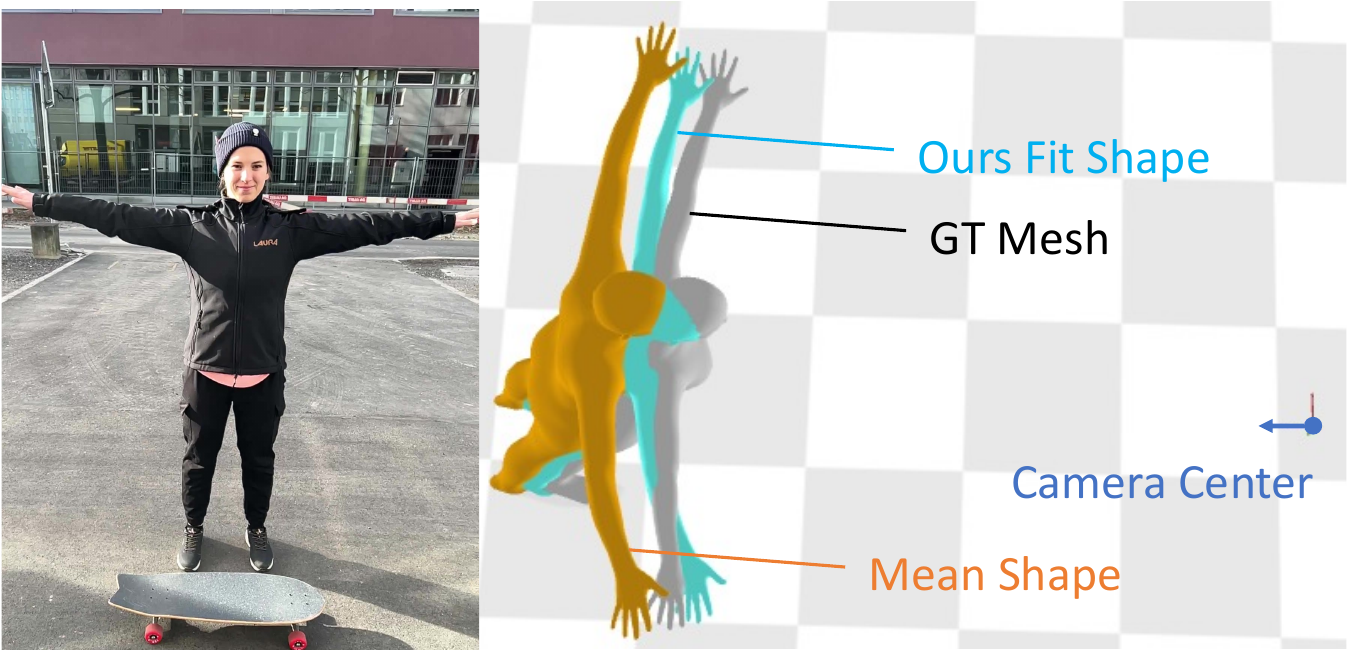}
\caption{\textbf{Effect of shape errors on pelvis accuracy}. The mean shape of ScoreHMR's predictions is less accurate, which leads to pelvis misalignments, as the prediction must match 2D reprojections. In contrast, our shape estimate (from SHAPify) is more accurate, allowing for a better estimation of pelvis position.}
\label{fig:shape_camera}
\end{figure}

In~\cref{fig:shape_camera}, we illustrate how an incorrect body shape adversely affects pelvis positioning. Using incorrect shape parameters, such as the mean shape, leads to using the wrong bone lengths for body fitting. This will not only affect the accuracy of the local pose but also the pelvis positions in the camera coordinate. Please also refer to our video for a better visualization of this effect.

\subsection{More qualitative results}
\begin{figure*}[th]
\centering
\includegraphics[width=\linewidth]{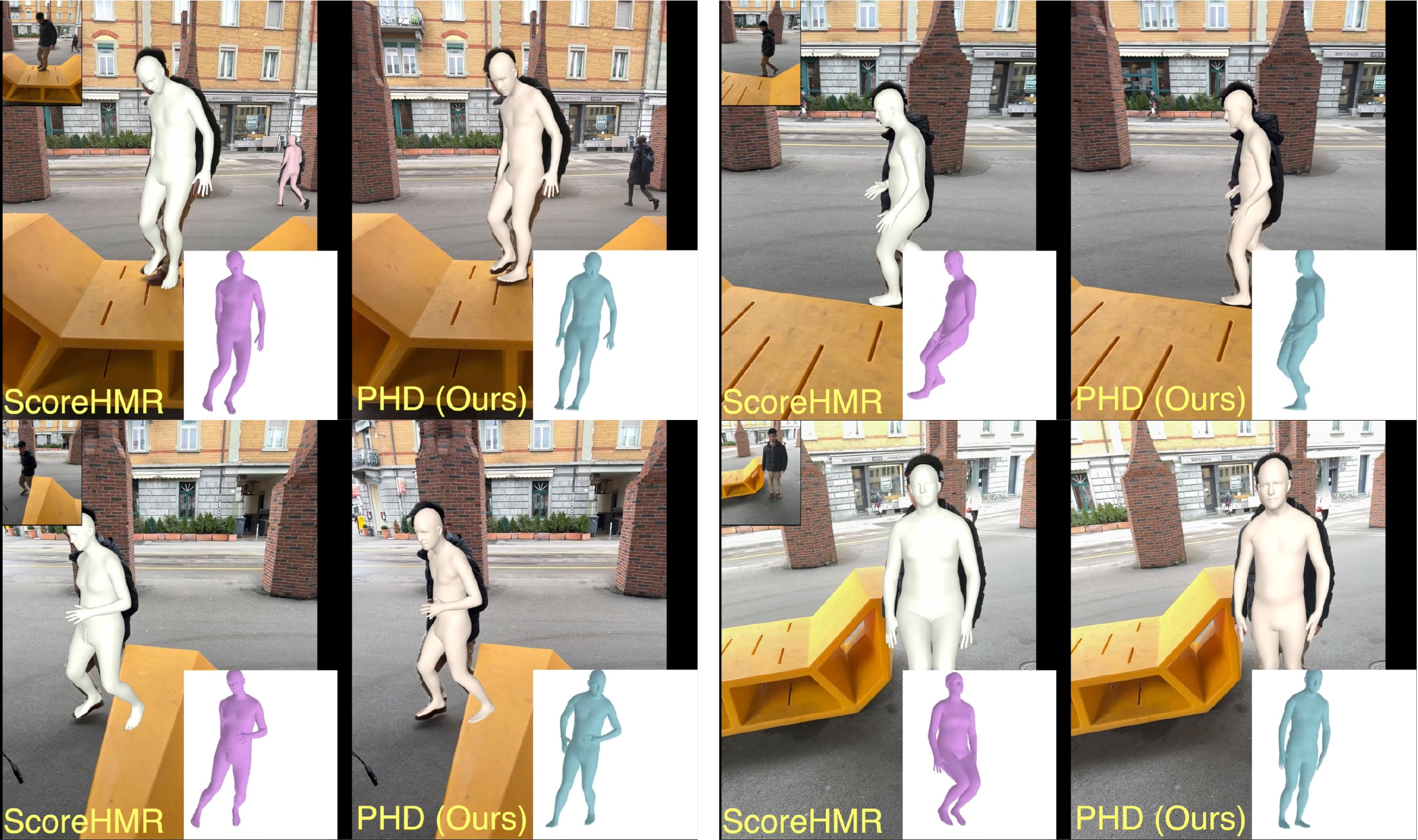}
\caption{\textbf{Quantitative comparisons on EMDB}. While the 2D reprojections appear to be correct, ScoreHMR often produces implausible body poses, such as bending knees and self-penetration. PHD addresses these issues with a stronger 3D pose prior, PointDiT.}
\label{fig:compare_vid}
\end{figure*}

\begin{figure*}[t]
\centering
\includegraphics[width=\linewidth]{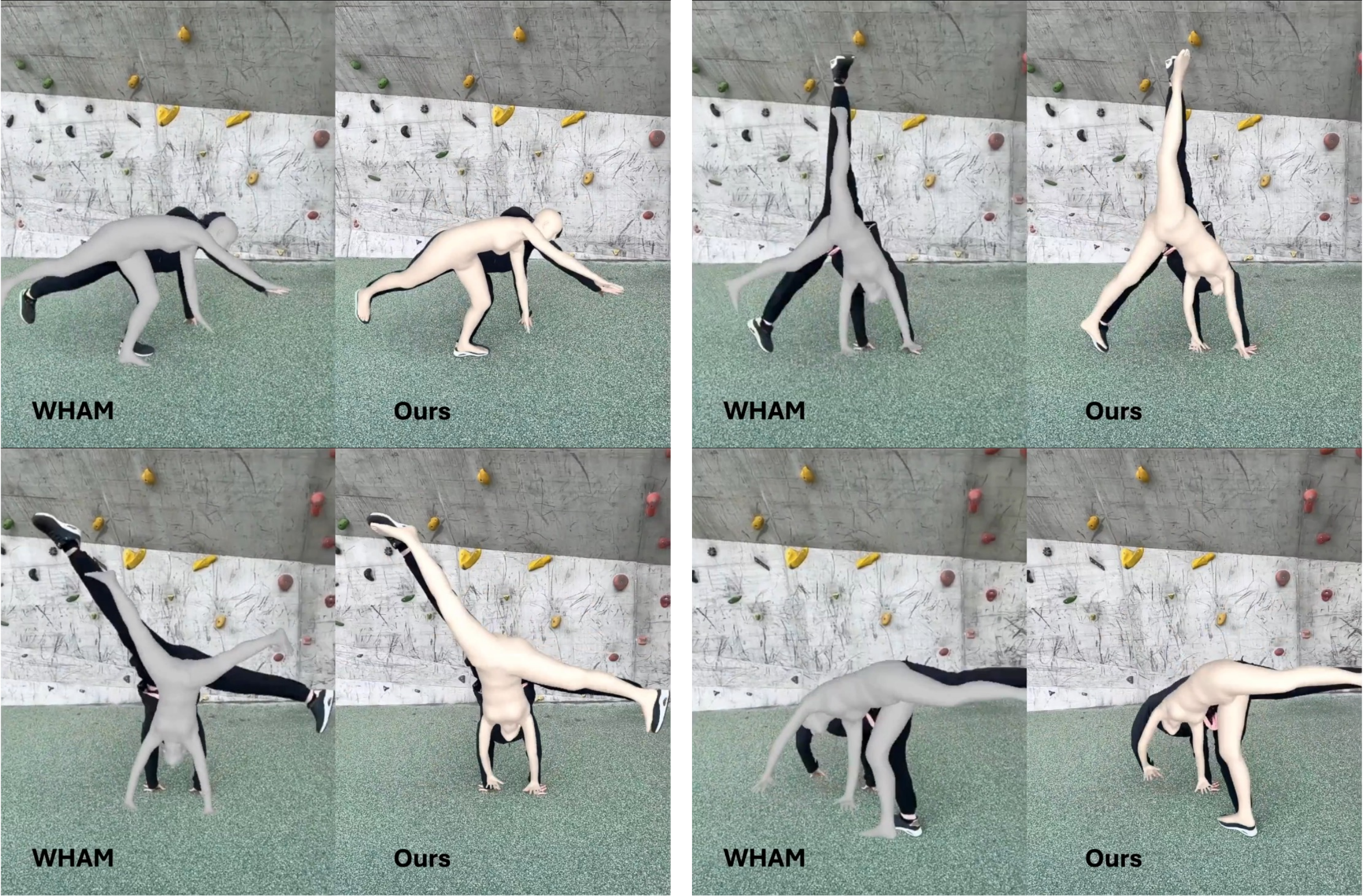}
\caption{\textbf{Quantitative comparisons on EMDB}. When comparing with the regression-based method, WHAM, our approach is more robust to challenging poses and achieves better 2D alignments.}
\label{fig:compare_wham}
\end{figure*}

We present more qualitative results with ScoreHRM~\cite{stathopoulos2024score} in~\cref{fig:compare_vid} and WHAM~\cite{shin2023wham} in~\cref{fig:compare_wham}. In~\cref{fig:wild}, we also showcase results of in-the-wild human performance capturing using our method with a modern smartphone.

\section{Discussion}

\subsection{Applications}
Our method is beneficial for in-the-wild avatar reconstruction~\cite{xu_monoperfcap,guo2023vid2avatar,guo2025vid2avatarpro,moon2024exavatar}. Existing avatar reconstruction methods all require accurate pose fitting as a starting point, with the most common practice being to use the averaged shape from a regressor and refine the poses using 2D keypoints or by jointly optimizing poses with appearance cues~\cite{lu2024avatarpose}. However, these strategies are only effective when pose errors are small and correctable. As discussed earlier, generalized regressors can easily produce implausible 3D poses, which prevent the avatar from learning accurate pose-dependent appearances.
Therefore, \methodname~offers a simple and effective solution by providing more accurate pose and shape inputs for in-the-wild avatar reconstruction. Furthermore, we believe that accurate personal shape information is a key factor for future avatar-centric applications, such as 3D virtual try-on and digital human interaction.

\begin{figure}[t]
\centering
\includegraphics[width=\columnwidth]{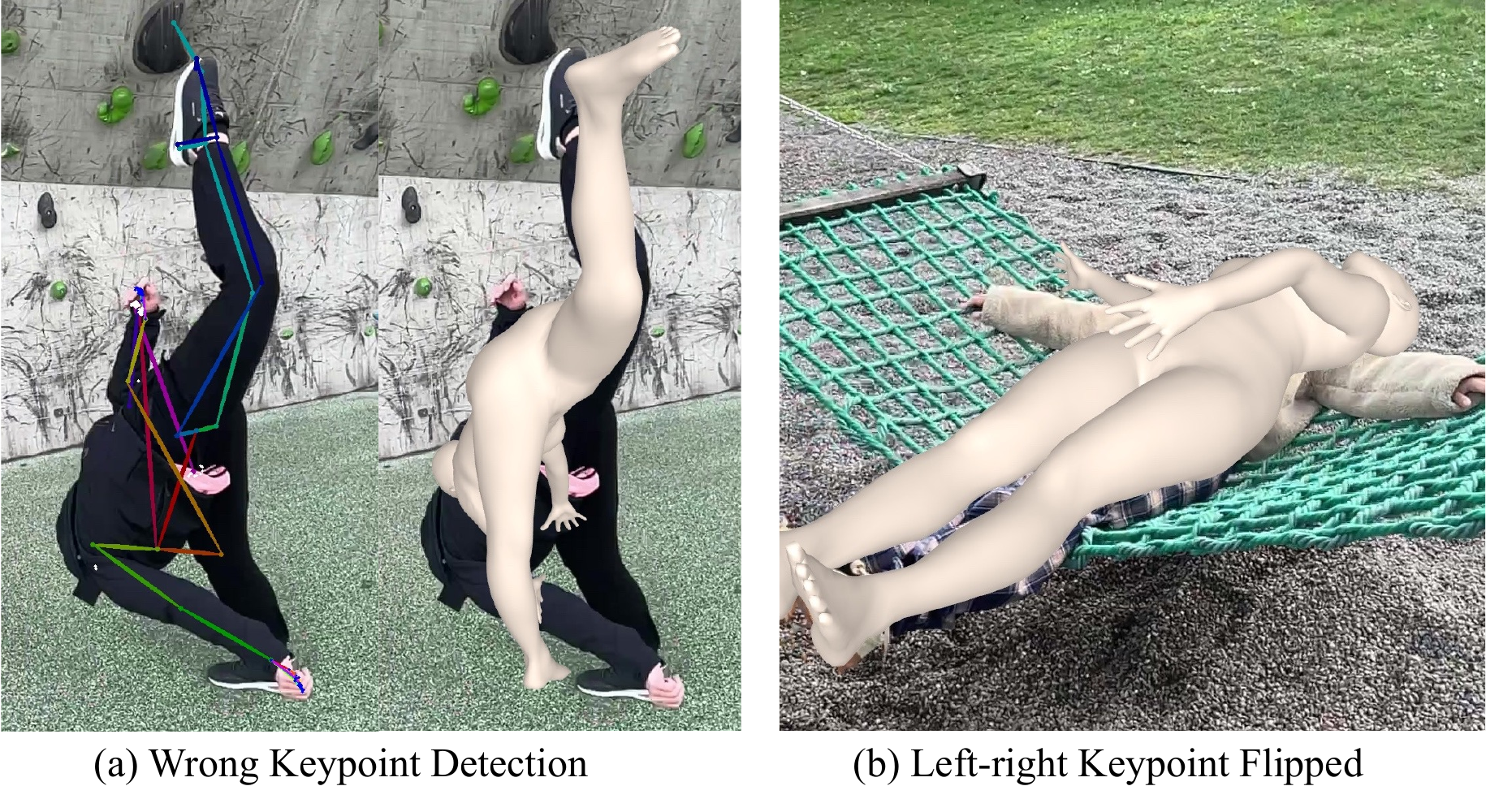}
\caption{\textbf{Failure cases of our method}. (a) Significantly wrong 2D keypoint detections that our pose prior cannot counteract. (b) The keypoint detector swapped left and right, which is difficult to recover from.}
\label{fig:fail}
\end{figure}

\subsection{Absolute Pose Metrics}

In~\cref{sec:compare_all}, we emphasized the importance of absolute pose accuracy in the camera coordinate system and showed that most state-of-the-art pose regressors failed to achieve desirable results. Recently, a new line of work~\cite{ye2023slahmr, yuan2022glamr, shin2023wham, jiang2024worldpose, shen2024gvhmr} has pursued a similar idea in a slightly different setup by estimating body poses and trajectories in the world coordinate system. Their evaluation metric, \textbf{G-MPJPE}, is determined by two factors: (1) how well the human poses and motion over time are estimated, and (2) how accurately the camera pose trajectories are recovered. Consequently, these approaches usually involve full-video SLAM tracking and global optimization over body and camera poses. This metric is more suitable for video-based methods.

In contrast, our proposed camera-coordinate metric, \textbf{C-MPJPE}, is better suited for per-frame or online methods that can be used for on-device computations (like ours). C-MPJPE disentangles the error caused by camera poses and evaluates only the accuracy of the body pose. Because personal shape (scale) information is fixed in our problem setting, C-MPJPE is also correlated with 2D alignment accuracy, as the 2D projection is definitive when the body scale is fixed. It is also worth noting that C-MPJPE can be converted to G-MPJPE by accounting for the camera extrinsics over time.

\subsection{Limitations and Future Work}

\paragraph{Hands and faces.}
Currently, our work focuses on fitting SMPL parameters to videos due to the limited availability of evaluation benchmarks for in-the-wild human pose estimation. It is worth noting that both our method and the PoseDiT model can be easily extended to any format of parametric models, such as SMPL-X. One of our future goals is to extend PointDiT to handle hand gestures and facial expressions. This can be readily achieved by selecting the corresponding data from the BEDLAM training set in SMPL-X format. We believe that for future perceptual AI systems, estimating holistic human body poses with precise hand and facial information will be indispensable. 

\paragraph{2D keypoint detection.}
Although PointDiT is an effective 3D prior, our fitting method still depends on the accuracy of 2D keypoint detections. When these detections fail significantly (see~\cref{fig:fail}), our method struggles to recover. An exciting direction for future work is to explore appearance- or feature-based fitting by leveraging powerful, pre-trained foundation models such as DINOv2~\cite{oquab2024dinov}. This approach could make the fitting process more robust to challenging poses that 2D keypoint detectors may fail.

\paragraph{Temporal Smoothness.}
Our method and the evaluations in this paper follow a per-frame setting. This setup is closer to real-world on-device computation, allowing for a fair assessment of the model's performance without relying on future or past information. While the metrics appear promising, we observed noticeable jittering across frames. Additionally, our method does not hallucinate continuous body motion when parts of the body are occluded or unobserved. To address these issues, a promising direction is to extend PointDiT into a temporal model that incorporates motion priors rather than predicting a single pose at a time. However, training such a model would require even more data beyond what BEDLAM provides. We believe that video generative AI can help address this challenge by offering a wide range of realistic motion data.

\paragraph{Learning-based optimization.}
Our method can run at 1 FPS on a common GPU (RTX 3090). While this is already faster than most existing pose fitting methods, it is still not yet suitable for real-time pose tracking. This limitation stems from the nature of optimization loops. Even though PointDiT provides strong 3D guidance during the fitting process, it still requires several iterations to converge. One way to mitigate these long optimization loops is to employ a "learning to optimize" approach~\cite{song2020humanbodymodelfitting,corona2022learned}, which enables the network to memorize optimization trajectories. Potentially, it achieves faster inference speeds through a feed-forward network while retaining the accuracy of optimization.

\clearpage

\end{document}